\pdfoutput=1

\documentclass[11pt]{article}

\usepackage[]{ACL2023}

\usepackage{times}
\usepackage{latexsym}
\usepackage{graphicx}
\usepackage{enumitem}
\usepackage{subfigure}
\usepackage{makecell}
\usepackage{amssymb}
\usepackage{lipsum}
\usepackage{booktabs}
\usepackage{multirow}
\usepackage{colortbl}
\usepackage{tabularx}
\usepackage{amsmath}
\usepackage{framed}
\usepackage{bbding}
\usepackage{lipsum}
\usepackage{longtable}
\usepackage{diagbox}
\usepackage[normalem]{ulem}
\useunder{\uline}{\ul}{}
\usepackage{CJKutf8}
\usepackage{fancyhdr}

\usepackage[utf8]{inputenc}
\usepackage{cleveref}
\crefname{section}{§}{§§}
\Crefname{section}{§}{§§}

\usepackage[T1]{fontenc}

\usepackage[utf8]{inputenc}

\usepackage{microtype}

\usepackage{inconsolata}

\title{Element-aware Summarization with Large Language Models: \\ Expert-aligned Evaluation and Chain-of-Thought Method}

\author{Yiming Wang, Zhuosheng Zhang, Rui Wang\thanks{\quad Corresponding author} \\
        Shanghai Jiao Tong University \\
        \texttt{alsaceym@gmail.com, \{zhangzs, wangrui12\}@sjtu.edu.cn}}

\begin{document}
\maketitle

\thispagestyle{fancy}

\begin{CJK*}{UTF8}{gbsn}

\begin{abstract}

Automatic summarization generates concise summaries that contain key ideas of source documents.
As the most mainstream datasets for the news sub-domain, \textit{CNN/DailyMail} and \textit{BBC XSum} have been widely used for performance benchmarking.
However, the reference summaries of those datasets turn out to be noisy, mainly in terms of factual hallucination and information redundancy.
To address this challenge, we first annotate new expert-writing \textbf{Element-aware} test sets following the ``Lasswell Communication Model'' proposed by \citet{lasswell1948structure}, allowing reference summaries to focus on more fine-grained news elements objectively and comprehensively.
Utilizing the new test sets, we observe the surprising zero-shot summary ability of LLMs, which addresses the issue of the inconsistent results between human preference and automatic evaluation metrics of LLMs' zero-shot summaries in prior work.
Further, we propose a \textbf{Summary Chain-of-Thought (SumCoT)} technique to elicit LLMs to generate summaries step by step, which helps them integrate more fine-grained details of source documents into the final summaries that correlate with the human writing mindset. 
Experimental results show our method outperforms state-of-the-art fine-tuned PLMs and zero-shot LLMs by +4.33/+4.77 in R{\footnotesize OUGE}-L on the two datasets, respectively.
Dataset and code are publicly available at \url{https://github.com/Alsace08/SumCoT}.

\end{abstract}

\section{Introduction}

\begin{figure}[tb]
  \centering
  \includegraphics[width=1.00\linewidth]{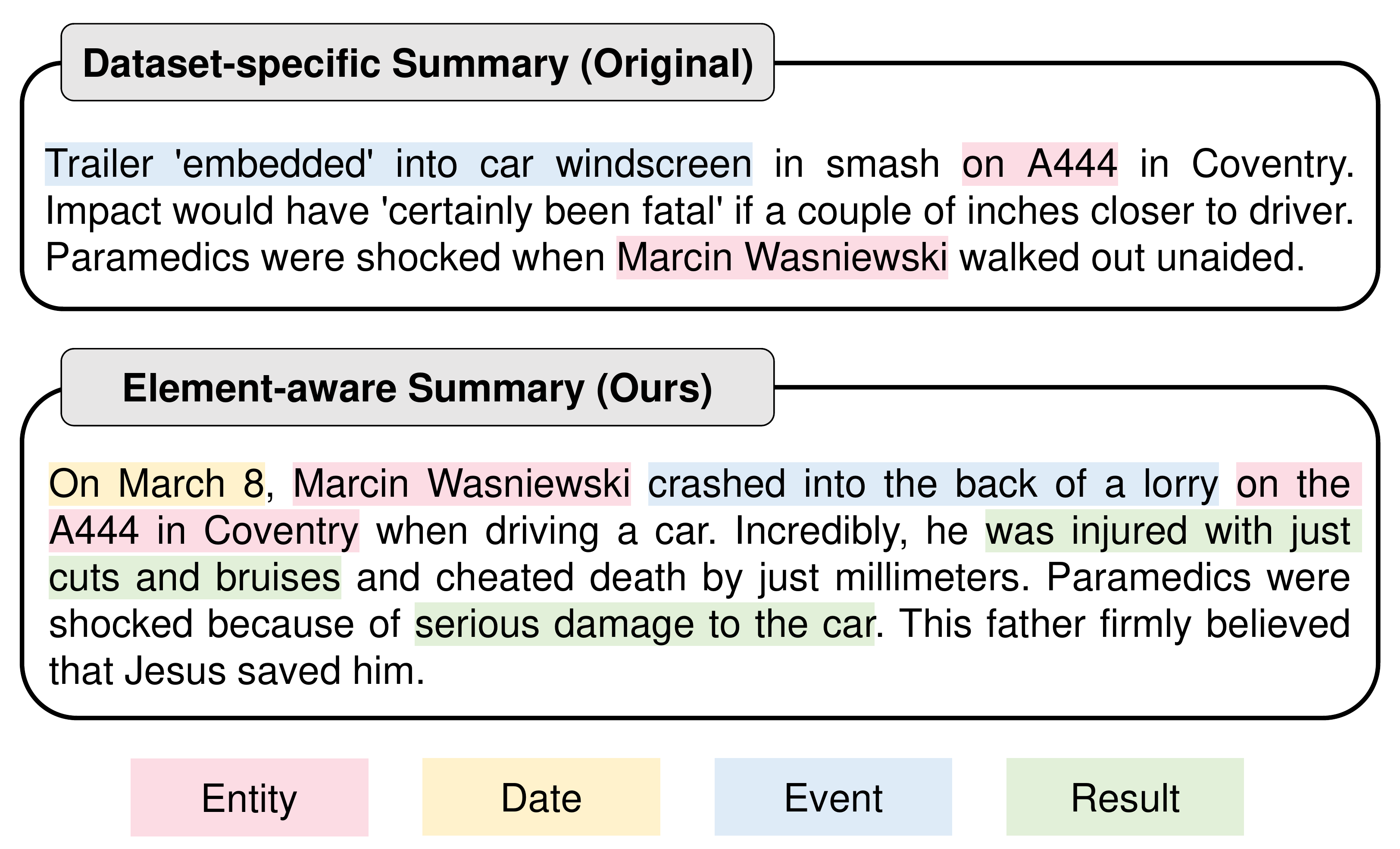}  
  \vspace{-0.25in}
  \caption{Case comparisons for \textbf{Element-aware summary} (ours) and dataset-specific summary (original). News elements have been highlighted with different color shadows. It is clear that our element-aware summary covers more comprehensive elements, and the logical connection between the elements is smoother.}
  \label{img:eg_first}
\end{figure}

\begin{figure*}[tb]
  \centering
  \includegraphics[width=1\linewidth]{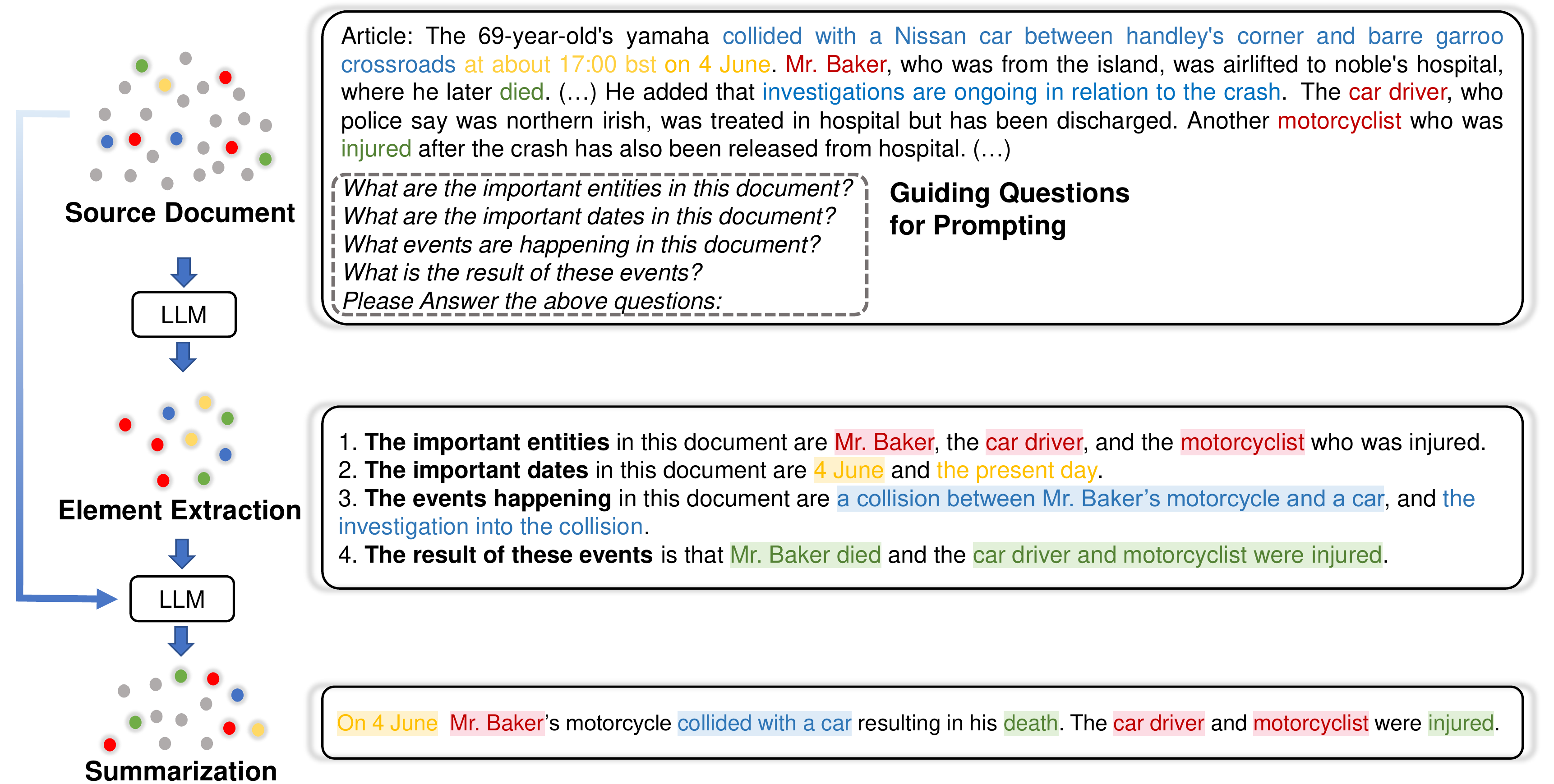}  
  \vspace{-0.25in}
  \caption{Full pipeline and example of our \textbf{Summary Chain-of-Thought} method.}
  \label{img:pipeline}
\end{figure*}

Automatic summarization is a challenging text generation task that condenses the source text into a few coherent and abstract sentences. In recent years, the study of summarization has evolved with supervised learning based on sequence-to-sequence architectures \cite{sutskever2014sequence,vinyals2015pointer,vaswani2017attention} and transfer learning based on pre-trained language models \cite{devlin2019bert,zhang2019hibert,liu2019roberta,lewis2020bart}.
Existing studies commonly train or fine-tune language models on large-scale corpus \cite{CNNDM,XSUM,wikihow,multinews}, so superior performance is often reported by measuring the lexical overlap (e.g. R{\footnotesize OUGE} \cite{lin2004rouge}) with golden summaries \cite{zhang2020pegasus,narayan2021planning,liu2022brio,narayan2022well}, which reflects the fit degree to these standard datasets.
However, some standard datasets have shown to be noise-enriched, mainly in terms of information redundancy \cite{kryscinski2019neural} and factual 
hallucination \cite{maynez2020faithfulness}. Meanwhile, sufficient experiments have shown that reference summaries in these standard datasets perform poorly on human assessment dimensions, especially coherence, consistency, and relevance \cite{stiennon2020learning,fabbri2021summeval}.

To fill this gap, this work releases expert-writing \textbf{Element-aware} summary test sets.
In professional news writing, core elements such as character, time, place, event, etc., are indispensable.
This theory named ``Lasswell Communication Model'' was first proposed by \citet{lasswell1948structure}, and later evolved into the ``5W1H'' paradigm.\footnote{\textit{who}, \textit{where}, \textit{when}, \textit{why}, \textit{what}, and \textit{how}. \textit{who} and \textit{where} can be packaged as \textit{entity}. \textit{why} is usually not independent of \textit{what}, so the two can be packaged as \textit{event}.}
Following this fine-grained protocol,\footnote{Some journalists may follow the Inverted Pyramid style \cite{po2003news}, but this protocol is more about a consideration of the full-text layout and is prone to information imbalance within the text \cite{wikihow}.} 
we ask three news experts to rewrite summaries of source documents from two standard news datasets --- \textit{CNN/DailyMail} \cite{CNNDM} and \textit{BBC XSum} \cite{XSUM}, allowing reference summaries to contain news core elements objectively and comprehensively\footnote{In the era of zero-shot paradigm, LLMs (e.g. GPT-3 \cite{gpt3}) have shown decent performance in summarization tasks, so this work focuses on the zero-shot setting to only annotate test sets.} (See Figure \ref{img:eg_first} for one example).
Utilizing the new test sets, we are surprised to find that the zero-shot performance of large language models (LLMs) is highly competitive with some strong fine-tuned pre-trained models (PLMs), and the performance of PLMs declines compared to standard test sets.
This observation can to some extent address the confusion raised by \citet{goyal2022news} that why GPT-3 generates more human-favored summaries but performs unexpectedly poorly in automatic evaluation metrics --- likely due to the limitation of noisy testing domains.

We further build a benchmark for the new test sets. Inspired by the competitive zero-shot performance of LLMs and chain-of-thought technique \cite{wei2022chain,kojima2022large}, we create \textbf{Sum}mary \textbf{C}hain-\textbf{o}f-\textbf{T}hought (\textbf{SumCoT}) to elicit LLMs to generate summaries step by step (shown in Figure \ref{img:pipeline}). Concretely,
we first guide LLMs to extract the four most core elements for standardized news texts --- \textit{Entity}, \textit{Date}, \textit{Event}, \textit{Result} --- through some manually-set guiding questions. Immediately after, the guiding questions and corresponding answers output by LLMs are packaged, they further guide LLMs to focus on more critical details to generate summaries that better correlate with the element-aware writing pattern.

Overall, our main contributions are three-fold:

(i) We construct expert-writing element-aware summary test sets to evaluate general summarization systems more objectively (§\ref{sec:testset}). 

(ii) We explore the zero-shot summarization ability of LLMs on the new test sets and demonstrate that their writing ability cannot be fully reflected by standard test sets (§\ref{sec:mainexperiments}). 

(iii)  We propose a new CoT-based summarization technique, which allows the LLMs to generate more fine-grained summaries step by step (§\ref{sec:sumcot}).

\section{Element-aware Summary Test Set}\label{sec:testset}

\subsection{Data Construction}\label{sec:dataset}

We select two standard news summary datasets (test sets) as document sources, which are representative in terms of length and abstraction:
\textbf{(i) \textit{CNN/DailyMail}} \cite{CNNDM} provides a large-scale multi-domain news collection, which is representative of single-document datasets. We use the standard splits \cite{hermann2015teaching} for test sets;
\textbf{(ii) \textit{BBC XSum}} \cite{XSUM} provides a highly abstracted news collection. It has one-sentence summaries and is more abstractive than the \textit{CNN/DailyMail} dataset.

For both datasets, we ask three news experts to independently write professional summaries for 200 randomly sampled source documents according to a complete writing protocol (introduced in §\ref{sec:protocol}), ensuring comprehensiveness, objectivity, and uniformity of writing style.
Different from crowd-sourcing, the involvement of professional writers allows higher inter-annotator agreement.
Also, to ensure the uniformity of writing style, we require one of the experts to lead the writing, and the other two to judge the completed summary in four dimensions from the protocol. 
If there exist inconsistent opinions, they will revise the summary after internal discussion until all pass this annotation.
Statistically, the annotation duration of one summary is approximately proportional to the length of source documents. For \textit{CNN/DailyMail}, a summary is written in 25-30 minutes on average, and for \textit{BBC XSum}, in 15-20 minutes on average.

\subsection{Writing Protocols}\label{sec:protocol}

Annotators must follow a comprehensive protocol when writing. 
Specifically, we divide the protocol into micro demands and macro demands. The former emphasizes our targets, namely element awareness, and the latter guarantees the professionalism and objectivity of the overall writing quality, which alleviates the simple stacking of elements. The two demands complement each other.

\paragraph{Micro Demands.}   All news summaries should have four essential core elements --- \textit{\textbf{Entity}}, \textit{\textbf{Date}}, \textit{\textbf{Event}}, and \textit{\textbf{Result}}
--- following the ``Lasswell Communication Model'' \cite{lasswell1948structure}, and these elements must be faithful to the source document. For example, when there is no date in the source document, writers can not add dates to the final summary by force.

\paragraph{Macro Demands.}  All news summaries should focus on four dimensions \cite{gehrmann2018bottom,kryscinski2019neural}.
\textbf{(i) Fluency}: No spelling, grammatical, or syntactic errors within sentences; \textbf{(ii) Coherence}: The summary should not be a heap of events, and linguistic transition must be smooth and logically correct; \textbf{(iii) Consistency}: No hallucinated facts --- neither facts that do not appear in or are contrary to the source document are allowed; \textbf{(iv) Relevance}: Adequately weigh the importance of multiple facts, and find the core concern of the text. 
Non-core facts can be reduced in length, and redundant details are not allowed.

\begin{table}[t]
\centering
\footnotesize
  \renewcommand\arraystretch{1}
  \setlength{\tabcolsep}{0.8mm}{
  \resizebox{1\columnwidth}{!}{
\begin{tabular}{l|cc}
\toprule

\multirow{3}{*}{\makecell[c]{\textbf{Reference}\\\textbf{ Summary}}} 
& \multicolumn{2}{c}{\textbf{\textit{CNN/DaliyMail}}} \\
\cmidrule(r){2-3} 

& \multirow{2}{*}{\makecell[c]{\% of novel\\\textbf{uni/bi/trigram}}}       
& \multirow{2}{*}{\makecell[c]{Avg. summary length of\\\textbf{words/sentences}}}
\\             
\\
\midrule

Dataset-specific & 17.00/53.91/71.98 & 50.14/3.59   \\ 
\rowcolor{gray!20}
Element-aware  & 20.31/49.72/62.14 & 51.08/2.71 \\

\midrule

\multirow{3}{*}{\makecell[c]{\textbf{Reference}\\\textbf{ Summary}}} 
& \multicolumn{2}{c}{\textbf{\textit{BBC XSum}}} \\
\cmidrule(r){2-3} 

& \multirow{2}{*}{\makecell[c]{\% of novel\\\textbf{uni/bi/trigram}}}       
& \multirow{2}{*}{\makecell[c]{Avg. summary length of\\\textbf{words/sentences}}}         
\\
\\
\midrule

Dataset-specific & 39.39/87.86/96.95 & 22.18/1.00      \\ 
\rowcolor{gray!20}
Element-aware  & 36.28/70.56/82.36 & 23.33/1.00 \\

\bottomrule

\end{tabular}}
    \caption{Some statistics of element-aware summaries compared with original dataset-specific summaries. \textit{Novel $n$-grams}
indicates the $n$-grams that are included in the summary but not in the source document.}
    \label{tab:stastical}%
}
    
\end{table}

\subsection{Overall Quality}\label{sec:quality}

We first compare the overall quality of our test sets with the original data.
Table \ref{tab:stastical} quantifies some statistics of the element-aware summaries compared with original dataset-specific summaries. The average length of element-aware summaries largely matches the distribution of that of dataset-specific summaries. In terms of abstraction, we report the percentage of novel $n$-grams
that are included in the summary but not in the source document. We note that the percent of novel $n$-grams in element-aware summaries is lower than that of dataset-specific summaries but with a reasonable gap, which reflects that expert-writing element-aware summaries would be more faithful to the source documents but not heavily replicate them.\footnote{Additionally, factual errors in the dataset-specific summaries will result in a spuriously high abstraction to some degree. In contrast, element-aware summaries better trade-off abstraction and faithfulness (See Appendix \ref{sec:abstraction} for examples).}

\begin{figure}
  \centering
    \subfigure[\textit{CNN/DailyMail}]{
        \begin{minipage}[b]{0.48\textwidth}
        \includegraphics[width=1\textwidth]{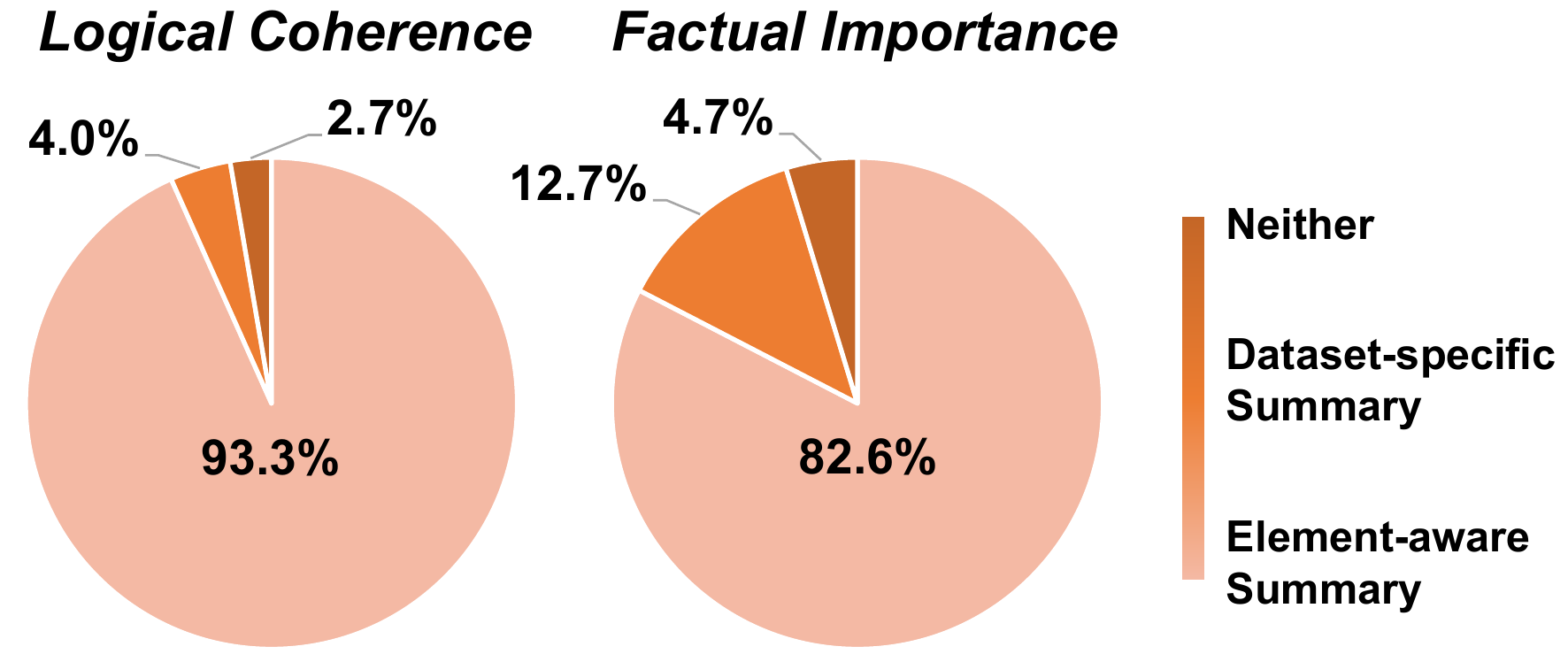}
        \end{minipage}
    \label{fig:case}
    }
    \subfigure[\textit{BBC XSum}]{
        \begin{minipage}[b]{0.48\textwidth}
        \includegraphics[width=1\textwidth]{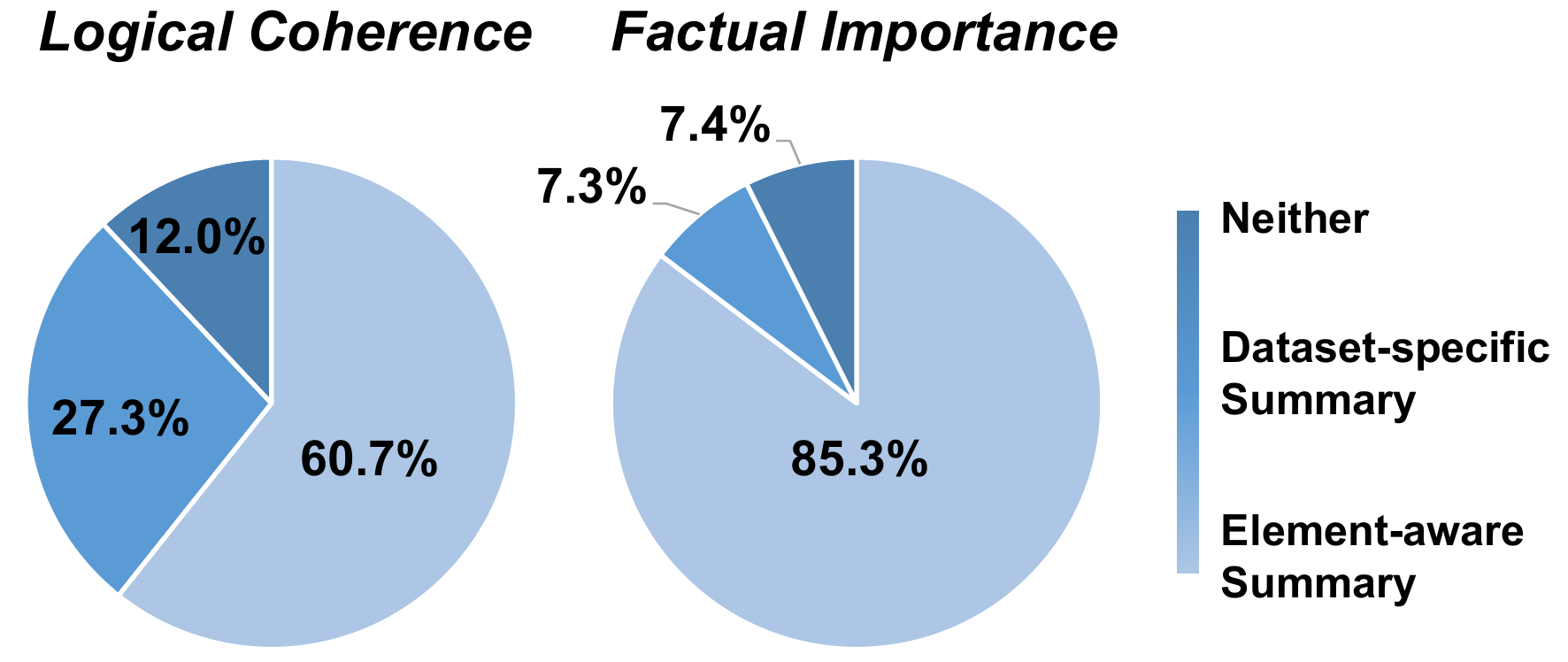}
        \end{minipage}
    \label{fig:pre-performance}
    	}
    \vspace{-0.1in}
    \caption{Average annotator vote distribution for better summaries between dataset-specific and element-aware summaries on ``logical coherence'' and ``factual importance'' dimensions. It is clear that element-aware summaries are more accepted by the public.}
    \label{img:data_analysis}
\end{figure}

We further hold a vote on two highly subjective dimensions --- logical coherence and factual importance, they reflect the professionalism and the information comprehensiveness of writing.\footnote{Whether the transition between facts is coherent, and whether important facts in the source documents are comprehensive and non-redundant in the summaries.}
We ask three annotators to perform preference selection on 50 randomly selected instances from both datasets — for each instance, they can select \textbf{at most one} summary that performs better in the two dimensions, respectively, or none if they consider both to be not good. 

Figure \ref{img:data_analysis} shows the vote results. It is clear that element-aware summaries are significantly more popularly accepted in both subjective dimensions by the public, demonstrating that our summaries are more human-favored.

\begin{table}[t]
\centering
\footnotesize
  \renewcommand\arraystretch{1}
  \setlength{\tabcolsep}{2.5mm}{
  \resizebox{0.98\columnwidth}{!}{
\begin{tabular}{l|ccc|ccc}
\toprule

\multicolumn{7}{c}{\textbf{\textit{CNN/DailyMail}}} \\
\midrule
\multirow{2}{*}{\makecell[l]{\textbf{Core}\\\textbf{Element}}} 
& \multicolumn{3}{c|}{\textbf{Element-aware}} 
& \multicolumn{3}{c}{\textbf{Dataset-specific}}\\
\cmidrule(r){2-7} 
& \textit{P} & \textit{R} & \textit{\textbf{F1}} & \textit{P} & \textit{R} & \textit{\textbf{F1}}\\      
\midrule

Entity & 0.98 & 0.96 & \textbf{0.98} & 0.75 & 0.63 & \textbf{0.68} \\
Date & 0.89 & 0.91 & \textbf{0.90} & 0.74 & 0.65 & \textbf{0.69}\\
Event & 0.96 & 0.95 & \textbf{0.95} & 0.66 & 0.55 & \textbf{0.60}\\
Result & 0.95 & 0.95 & \textbf{0.95} & 0.49 & 0.42 & \textbf{0.45}\\

\midrule

\multicolumn{7}{c}{\textbf{\textit{BBC XSum}}} \\
\midrule
\multirow{2}{*}{\makecell[l]{\textbf{Core}\\\textbf{Element}}} 
& \multicolumn{3}{c|}{\textbf{Element-aware}} 
& \multicolumn{3}{c}{\textbf{Dataset-specific}}\\
\cmidrule(r){2-7} 
& \textit{P} & \textit{R} & \textit{\textbf{F1}} & \textit{P} & \textit{R} & \textit{\textbf{F1}}\\    
\midrule

Entity & 0.97 & 0.87 & \textbf{0.92} & 0.76 & 0.54 & \textbf{0.63}\\
Date & 0.97 & 0.95 & \textbf{0.96} & 0.52 & 0.45 & \textbf{0.48}\\
Event & 0.93 & 0.93 & \textbf{0.93} & 0.80 & 0.48 & \textbf{0.60}\\
Result & 0.96 & 0.98 & \textbf{0.97} & 0.23 & 0.18 & \textbf{0.20}\\

\bottomrule                     
        
\end{tabular}}
    \caption{The comparison between element-aware and dataset-specific test sets over $\mathrm{Precision}$ (\textit{P}), $\mathrm{Recall}$ (\textit{R}), and $F_1$ score of all four elements.}
    \label{tab:elementaware}%
}

\end{table}

\begin{table*}[t]
\centering
\footnotesize
  \renewcommand\arraystretch{1}
  \setlength{\tabcolsep}{2.5mm}{
  \resizebox{2\columnwidth}{!}{
\begin{tabular}{l|ccc|c|ccc|c}
\toprule

\multicolumn{9}{c}{\textbf{\textit{CNN/DaliyMail}}} \\
\midrule

\multirow{2}{*}{\diagbox{\textit{Model}}{\textit{Ref}}} & \multicolumn{4}{c|}{\textbf{Element-aware (ours)}} & 
\multicolumn{4}{c}{\textbf{Dataset-specific (original)}} \\
\cmidrule(r){2-9} 

& \multicolumn{1}{c|}{\textbf{R{\scriptsize OUGE}-1}}       
& \multicolumn{1}{c|}{\textbf{R{\scriptsize OUGE}-2}}       
& \multicolumn{1}{c|}{\textbf{R{\scriptsize OUGE}-L}}
& \textbf{BERTS{\scriptsize CORE}}
& \multicolumn{1}{c|}{\textbf{R{\scriptsize OUGE}-1}}       
& \multicolumn{1}{c|}{\textbf{R{\scriptsize OUGE}-2}}       
& \multicolumn{1}{c|}{\textbf{R{\scriptsize OUGE}-L}}
& \multicolumn{1}{c}{\textbf{BERTS{\scriptsize CORE}}}
\\                        
\midrule

B{\scriptsize ART}-B{\scriptsize ASE} & 36.06 & 15.93 & 33.09 & 0.8762 & 38.55 & 17.57 & 36.05 & 0.8779\\
B{\scriptsize ART}-L{\scriptsize ARGE} & \textbf{37.98} & \textbf{18.16} & 34.30 & 0.8860 & 39.01 & 18.26 & \textbf{37.15} & \textbf{0.8868} \\
T5-L{\scriptsize ARGE} & 37.47 & 17.66 & \textbf{34.34} & 0.8768 & 38.84 & \textbf{18.39} & 37.01 & 0.8802\\
P{\scriptsize EGASUS}-L{\scriptsize ARGE} & 36.65 & 17.58 & 33.84 & 0.8710 & \textbf{39.11} & 17.82 & 36.86 & 0.8798\\

\midrule

175B GPT-3 & 37.75 & 15.20 & 34.25 & \textbf{0.8905} & 30.10 & 8.98 & 27.51 & 0.8718 \\

\midrule

\multicolumn{9}{c}{\textbf{\textit{BBC XSum}}} \\
\midrule

\multirow{2}{*}{\diagbox{\textit{Model}}{\textit{Ref}}} & \multicolumn{4}{c|}{\textbf{Element-aware (ours)}} & 
\multicolumn{4}{c}{\textbf{Dataset-specific (original)}} \\
\cmidrule(r){2-9} 

& \multicolumn{1}{c|}{\textbf{R{\scriptsize OUGE}-1}}       
& \multicolumn{1}{c|}{\textbf{R{\scriptsize OUGE}-2}}       
& \multicolumn{1}{c|}{\textbf{R{\scriptsize OUGE}-L}}
& \textbf{BERTS{\scriptsize CORE}}
& \multicolumn{1}{c|}{\textbf{R{\scriptsize OUGE}-1}}       
& \multicolumn{1}{c|}{\textbf{R{\scriptsize OUGE}-2}}       
& \multicolumn{1}{c|}{\textbf{R{\scriptsize OUGE}-L}}
& \multicolumn{1}{c}{\textbf{BERTS{\scriptsize CORE}}}
\\                        
\midrule

B{\scriptsize ART}-B{\scriptsize ASE} & 21.89 & 5.13 & 17.19 & 0.8663 & 29.67 & 10.09 & 24.46 & 0.8779\\
B{\scriptsize ART}-L{\scriptsize ARGE} & 23.79 & 5.02 & 17.93 & 0.8710 & 33.95 & 11.29 & 26.78 & 0.8880\\
T5-L{\scriptsize ARGE} &  24.98 & 6.89 & 19.46 & 0.8728 & 30.79 & 9.61 & 24.73 & 0.8792\\
P{\scriptsize EGASUS}-L{\scriptsize ARGE} & 21.35 & 4.87 & 17.03 & 0.8662 & \textbf{35.16} & \textbf{13.21} & \textbf{29.30} & \textbf{0.8888}\\

\midrule

175B GPT-3 & \textbf{31.74} & \textbf{10.95} & \textbf{25.42} & \textbf{0.8933} & 19.99 & 3.69 & 15.86 & 0.8654\\ 

\bottomrule                   
        
\end{tabular}}
    \caption{Performance comparison of zero-shot LLMs (175B GPT-3) and fine-tuned PLMs (B{\footnotesize ART}, T5, and P{\footnotesize EGASU}). We separately compare generated summaries of these models with original reference summaries from standard datasets (Dataset-specific) and our reference summaries rewritten by news experts (Element-aware). Results are evaluated automatically over R{\footnotesize OUGE}-1/2/L and BERTS{\footnotesize CORE}.}
    \label{tab:mainresults}%
}
    
\end{table*}

\subsection{Element-aware Characteristic}\label{sec:elementaware}

In this part, we will demonstrate that 
our annotated summaries have more obvious element-aware characteristic than the dataset-specific summaries. 

We ask three annotators to evaluate every document-summary pair.
For each sample, and for $i$-th annotator $(i=1,2,3)$ and $j$-th element in the writing protocol $(j=1,2,3,4)$, we ask this annotator to release two sets that separately contain all $j$-th elements in the source document they consider important and all $j$-th elements appearing in the summary.
The annotator-released sets for the source document and summary are denoted as $A_i^j$ and ${A'_i}^j$, respectively.

Then, we compute the $\mathrm{Precision}$ and $\mathrm{Recall}$, they separately reflect the accuracy of the core elements embedded in the summary and the hit rate of the core elements in the source document.
$\mathrm{Precision}^j$ and $\mathrm{Recall}^j$ are formulated as:\footnote{In extreme situations, when $A_i^j$ is empty, i.e., the annotator thinks that there is no $j$-th element in the source document, the Recall$^j$ is 1 if this element is also not covered in the summary, otherwise 0. Ditto for Precision$^j$ when ${A'_i}^j$ is empty.}

{
\begin{equation}
\begin{aligned}
    \mathrm{Precision}^j &= \frac{1}{3} \sum_{i=1}^3 \frac{|A_i^j \bigcap {A'_i}^j|}{|{A'_i}^j|}, \quad j = 1,2,3,4 \\
    \mathrm{Recall}^j &= \frac{1}{3} \sum_{i=1}^3 \frac{|A_i^j \bigcap {A'_i}^j|}{|A_i^j|}, \quad j = 1,2,3,4 \\
\end{aligned}
\end{equation}
}%
where $|\cdot|$ denotes the number of elements in the set. For \textit{\textbf{Event}} and \textit{\textbf{Result}}, a complete lexical overlap is unrealistic due to the subjectivity in expression, so as long as the same meaning is considered correct.

We compare the $\mathrm{Precision}$ and $\mathrm{Recall}$ between element-aware and dataset-specific test sets, and computer the average of all document-summary pairs of a test set.
We also compute $F_1$ score (The harmonic mean of $\mathrm{Precision}$ and $\mathrm{Recall}$) to measure the overall level.
Results are shown in Table \ref{tab:elementaware}, the comparison shows that our test sets have a significant advantage in the element-aware characteristic. The dataset-specific test sets perform poorly particularly in the $\mathrm{Recall}$ score, meaning that they have ignored many fine-grained details.


\section{Preliminary Comparison: Zero-shot LLMs Versus Fine-tuned PLMs}\label{sec:mainexperiments}

In this section, we preliminarily compare existing strong LLMs and PLMs upon our element-aware test sets, designed to analyze the general summary capabilities of zero-shot LLMs and fine-tuned PLMs from a more fine-grained perspective.

\subsection{Experimental Setup}
\label{sec:setup}

\paragraph{Dataset.}  We perform experiments on two mainstream news datasets \textit{CNN/DailyMail} and \textit{BBC XSum} introduced in §\ref{sec:dataset}. 
For each source document on both datasets, we compare summaries generated by models with dataset-specific (original) and element-aware (ours) reference summaries. Each test set includes 200 document-summary pairs consistent with the annotation number.

\paragraph{Models.}  For LLMs, We use 175B-parameter GPT-3 (\texttt{text-davinci-002} version) \cite{gpt3,ouyang2022training} for our study. For PLMs, we select 
B{\footnotesize ART} \cite{lewis2020bart}, 
T5 \cite{2020t5} --- two strong generation-oriented PLMs, and P{\footnotesize EGASUS} \cite{zhang2020pegasus} --- a summarization-customized PLM
fine-tuned on two datasets separately as the strong baselines.

\paragraph{Implementation.}  We follow the official fine-tuned models released on the Huggingface for PLMs generation.
For zero-shot prompts of LLMs, We follow \citet{sanh2022multitask} and \citet{goyal2022news} to set $\mathsf{[p]}$ = "\textit{Summarize the above article:}" as the standard prompt on \textit{CNN/DailyMail}. On \textit{BBC XSum}, considering its one-sentence summary style with extreme generalization, we use sentence-constraint prompt $\mathsf{[p]}$ = "\textit{Summarize the above article in one sentence:}". All the source documents are truncated to 1024 tokens when using PLMs and 2048 tokens when using LLMs. See Appendix \ref{sec:details} for more useful implementation details.

\paragraph{Evaluation.}  We evaluate the generated summaries using lexical-overlap metrics, specifically R{\footnotesize OUGE}-1/2/L \cite{lin2004rouge}, and embedding-similarity metrics, specifically BERTS{\footnotesize CORE} \cite{zhang2019bertscore}. Besides, we resort to more precise human studies to evaluate the consistency of generated summaries and source documents. See Appendix \ref{sec:details} for more useful evaluation details.

\subsection{Main Results}

\paragraph{Longitudinal Comparison: Language Models.}  First, we compare the performance of different models on the same test set (see columns of Table \ref{tab:mainresults}).
On dataset-specific test sets (the right part), the relative performances among PLMs are basically in line with the experimental results in \citet{zhang2020pegasus}, meaning that our sampled source documents basically follow the distribution of original test sets.
On element-aware test sets (the left part), surprisingly, zero-shot GPT-3 performs competitively with all other fine-tuned PLMs and even outperforms other models with a wide margin on \textit{BBC XSum}.
These all present that LLMs have more fine-grained summary capabilities, and their zero-shot evaluation is limited by the original test sets.

\paragraph{Horizontal Comparison: Test Sets.}
Next, we compare the performances of the same model on different test sets (see rows of Table \ref{tab:mainresults}).
We note that these fine-tuned PLMs perform worse on element-aware test sets than they do on dataset-specific test sets, with a particularly salient drop on \textit{BBC XSum}.
In contrast, GPT-3 obtains dramatic improvements on element-aware test sets. Compared with the performances on dataset-specific test sets, R{\footnotesize OUGE}-1/2/L increases by +7.65/+6.22/+6.74 points on \textit{CNN/DailyMail} and +11.75/+7.26/+9.56 points on \textit{BBC XSum}.
These contrasting results demonstrate that our annotated test sets pose a challenge for PLMs fine-tuned with standard datasets, but LLMs can perform well due to their more fine-grained writing capabilities.

\subsection{Human Study} 
\label{sec:human}
Human studies are conducted as an overall quality assessment of human preferences.
We use a 7-point Likert scale \cite{likert1932technique} to ask annotators to evaluate four dimensions: Fluency, Coherence, Consistency, and Relevance (equivalent to macro demands in §\ref{sec:protocol}).
Different from baseline-free human studies, we set the element-aware summaries as the baseline (score 0) and set the scoring range to -3\textasciitilde 3. A more positive score means higher quality than the element-aware summary and vice versa. 
For each sample, we present the dataset-specific (original), B{\footnotesize ART}-L{\footnotesize ARGE}, T5-L{\footnotesize ARGE}, P{\footnotesize EGASU}-L{\footnotesize ARGE} and 175B GPT-3 summaries to the annotators and ask them to score one by one.

\begin{figure}[tb]
  \centering
  \includegraphics[width=1.00\linewidth]{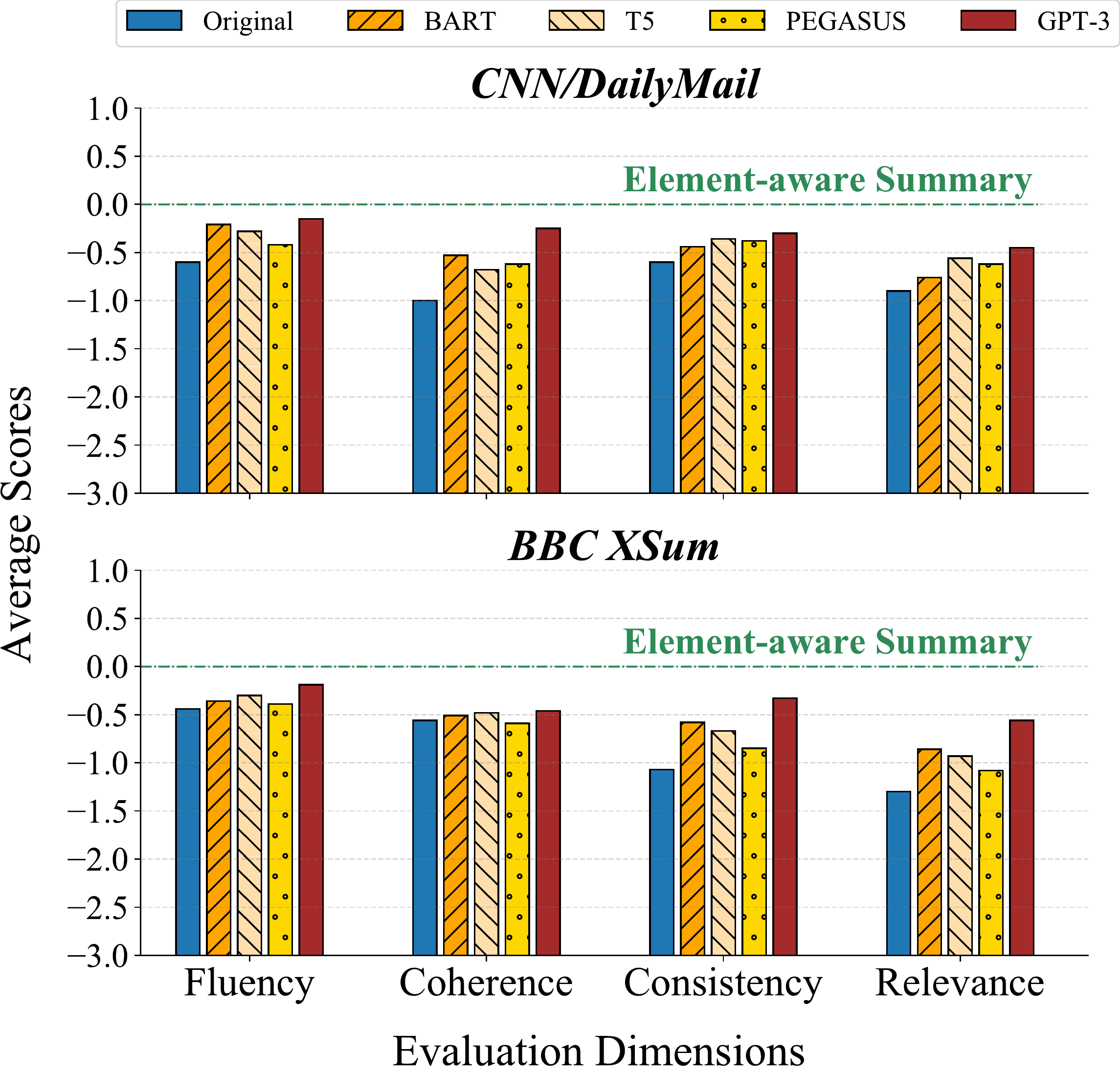}  
  \vspace{-0.25in}
  \caption{Human evaluation scores of four dimensions about summary quality on the 50-shot \textit{CNN/DailyMail} (the upper part) and \textit{BBC XSum} (the lower part) datasets. More human study details are shown in Appendix \ref{human_study_setup}.}
  \label{img:human}
\end{figure}

As is shown in Figure \ref{img:human}, GPT-3 summaries outperform almost all other dataset-specific or model-generated summaries in each dimension, although not yet achieved the level of element-aware summaries. 
All of these results can fully demonstrate that LLMs have great potential for summarization, and a higher-quality dataset is key for evaluation.


\begin{table*}[t]
\centering
\footnotesize
  \renewcommand\arraystretch{1}
  \setlength{\tabcolsep}{2.5mm}{

\begin{tabular}{p{1.98\columnwidth}}

\toprule

\textbf{GPT-3 Zero-shot Summary (Standard)} \\
\midrule
\textit{Summarize the above article:} \\
The article discusses how \textcolor{red}{Rod Blagojevich}, a former Illinois governor, has let his hair go while \textcolor{blue}{serving his prison sentence}. The photos show the former governor with white hair, rather than the black hair that was his trademark as a politician. Blagojevich is currently serving \textcolor{green}{a 14-year sentence} at \textcolor{red}{the Federal Correctional Institution Englewood} near Denver. \\

\midrule
\textbf{GPT-3 Zero-shot-cot Summary (Using SumCoT)} \\
\midrule
\textit{Let’s integrate the above information and summarize the article:} \\
In \textcolor{yellow}{2008}, \textcolor{red}{Rod Blagojevich} \textcolor{blue}{was arrested} for trying to sell Barack Obama's U.S. Senate seat. He was impeached and removed from office in \textcolor{yellow}{2009}, and \textcolor{green}{sentenced to 14 years in prison}. He is currently \textcolor{blue}{serving his sentence} at \textcolor{red}{the Federal Correctional Institution Englewood near Denver}. In \textcolor{yellow}{2014}, his lawyers attempted to \textcolor{blue}{appeal his conviction}, arguing that a recent Supreme Court ruling supported their view that Blagojevich was engaged in legal, run-of-the-mill political horse trading, not corruption.  \\

\bottomrule

\end{tabular}
}

\caption{Case comparisons between GPT-3 zero-shot summaries before and after using SumCoT. Spans of \textbf{Entity}, \textbf{Date}, \textbf{Event} and \textbf{Result} are separately highlighted in \textcolor{red}{red}, \textcolor{yellow}{yellow}, \textcolor{blue}{blue} and \textcolor{green}{green}. \textbf{Prompts} are presented in \textit{italics}.}
\label{tab:one_eg}%

\end{table*}


\begin{table*}[t]
\centering
\small
  \renewcommand\arraystretch{1}
  \setlength{\tabcolsep}{4.5mm}{
  {
\begin{tabular}{l|ccc|c}
\toprule

\multirow{2}{*}{Model} 
& \multicolumn{4}{c}{\textbf{\textit{CNN/DaliyMail}}} \\
\cmidrule(r){2-5}

& \multicolumn{1}{c|}{\textbf{R{\scriptsize OUGE}-1}}       
& \multicolumn{1}{c|}{\textbf{R{\scriptsize OUGE}-2}}       
& \multicolumn{1}{c|}{\textbf{R{\scriptsize OUGE}-L}}
& \multicolumn{1}{c}{\textbf{BERTS{\scriptsize CORE}}}
\\                        
\midrule

Previous SOTA in Table \ref{tab:mainresults} & 37.98 & 18.16 & 34.34 & 0.8905 \\
175B GPT-3 & 37.75 & 15.20 & 34.25 & 0.8905 \\

\rowcolor{gray!20}
175B GPT-3 w/ SumCoT & \textbf{43.03} ($\uparrow$  \textbf{5.05}) & \textbf{19.51} ($\uparrow$ \textbf{1.35}) & \textbf{38.67} ($\uparrow$  \textbf{4.33}) & \textbf{0.9023} ($\uparrow$  \textbf{0.0118}) \\

\midrule

\multirow{2}{*}{Model} 
& \multicolumn{4}{c}{\textbf{\textit{BBC XSum}}} \\
\cmidrule(r){2-5}

& \multicolumn{1}{c|}{\textbf{R{\scriptsize OUGE}-1}}       
& \multicolumn{1}{c|}{\textbf{R{\scriptsize OUGE}-2}}       
& \multicolumn{1}{c|}{\textbf{R{\scriptsize OUGE}-L}}
& \multicolumn{1}{c}{\textbf{BERTS{\scriptsize CORE}}}
\\                        
\midrule

Previous SOTA in Table \ref{tab:mainresults} & 31.74 & 10.95 & 25.42 & 0.8933 \\
175B GPT-3 & 31.74 & 10.95 & 25.42 & 0.8933 \\

\rowcolor{gray!20}
175B GPT-3 w/ SumCoT & \textbf{35.70} ($\uparrow$  \textbf{3.96}) & \textbf{15.31} ($\uparrow$  \textbf{4.36}) & \textbf{30.19} ($\uparrow$  \textbf{4.77}) & \textbf{0.9018} ($\uparrow$  \textbf{0.0085}) \\

\bottomrule                   
        
\end{tabular}
}
    \caption{Performance comparisons upon element-aware test sets of our method (GPT-3 with SumCoT), standard GPT-3, and previous state-of-the-art (SOTA) results in Table \ref{tab:mainresults} over each metric. The $\boldsymbol{\uparrow}$ and corresponding numbers on the right of each result of our method represent the increase after comparing with the previous SOTA.}
    \label{tab:sumcot}%
}
    
\end{table*}

\section{Towards Element-oriented Summary: Chain-of-Thought Method}\label{sec:sumcot}

We have analyzed the summary writing ability of zero-shot GPT-3 and other fine-tuned PLMs in §\ref{sec:mainexperiments}. 
We see that GPT-3 performs surprisingly well on our element-aware test sets. The results compellingly show that GPT-3 has great potential for fine-grained zero-shot summary writing. Inspired by the prevalence of the chain-of-thought (CoT) technique in LLMs \cite{wei2022chain,kojima2022large}, we can further enhance the summarization ability of LLMs by leveraging a CoT-based method (SumCoT). SumCoT elicits LLMs to focus on news core elements, thereby generating element-aware summaries step by step.
The pipeline and example have been illustrated in Figure \ref{img:pipeline}.

\begin{table}[tbp]
\centering
\footnotesize
\resizebox{\columnwidth}{!}
{
\begin{tabular}{l|c|c}
\toprule

\multirow{2}{*}{\textbf{Model}} 
& \textit{\textbf{CNN/DailyMail}} & \textit{\textbf{BBC XSum}} \\

& \textit{Flu/Coh/Con/Rel} & \textit{Flu/Coh/Con/Rel} \\
\midrule

175B GPT-3  & -0.18/-0.33/-0.37/-0.72 & -0.19/-0.48/-0.33/-0.56 \\

\rowcolor{gray!20}
\quad w/ SumCoT & \textbf{-0.10/-0.05/-0.23/-0.28} & \textbf{-0.11/-0.19/-0.07/-0.22}\\

\bottomrule
\end{tabular}}

\vspace{-0.1in}
\caption{Human evaluation scores (Scale -3\textasciitilde 3, and 0 represents the level of element-aware summaries) for zero-shot summaries of GPT-3 w/o and w/ SumCoT. \textit{Flu/Coh/Con/Rel} stands for Fluency/Coherence/Consistency/Relevance respectively.}
\label{tab:human_study_cot}%
\end{table}%

\subsection{Two-stage Pipeline}

We first ask the LLMs to extract core news elements in the source document by manually-set guiding questions, and later integrate the information based on the extracted elements and more details from the source documents. Pipeline details are as follows.

\begin{itemize}[leftmargin=*]

\item \textbf{Stage 1: Core element extraction.}  In the first stage, we create guiding-question prompts to elicit the LLMs to extract four core elements: \textbf{\textit{Entity}}, \textbf{\textit{Date}}, \textbf{\textit{Event}}, \textbf{\textit{Result}}.
For the $i$-th element, we set a simple question $q_i$ to guide the model for extracting it (shown in Figure \ref{img:pipeline}), and then concatenate these questions into $\boldsymbol{Q}=[q_1,q_2,q_3,q_4]$. Let the source document be $\mathbf{S}$, then the LLMs input in this stage is formulated as $[\boldsymbol{S};\boldsymbol{Q}]$.

\vspace{-0.1in}
\item  \textbf{Stage 2: Multiple information integration and summarization.} We obtain an extraction answer $\boldsymbol{A}$ from the LLMs in Stage 1.
Next, we integrate the extracted elements and more detailed information from the source document. We concatenate the source document, questions, answer, and a simple prompt $\mathsf{[p']}$="\textit{Let's integrate the above information and summarize the article:}" to prompt the LLMs for summary generation.\footnote{Similarly, for XSum, $\mathsf{[p']}$="\textit{Let's integrate the above information and summarize the article in one sentence:}"}
The input in this stage is formulated as $[\boldsymbol{S};\boldsymbol{Q};\boldsymbol{A};\mathsf{[p']}]$, and the output is the final summary.

\end{itemize}

\subsection{Comprehensive Evaluation}

First, we visually compare the quality of summaries generated by GPT-3 before and after using SumCoT. As shown in Table \ref{tab:one_eg}, it is clear that the summary generated under SumCoT contains more abundant fine-grained elements, saturating the summary text with more key information.

Next, we perform quantitative evaluations over the same metrics as in §\ref{sec:setup}.
We mainly compare our method (GPT-3 with SumCoT), standard GPT-3, and previous state-of-the-art (SOTA) results in Table \ref{tab:mainresults}, and updated results are shown in Table \ref{tab:sumcot}.
Compared with the standard GPT-3 and previous SOTA, GPT-3 with SumCoT obtains salient improvement in all metrics when compared with the element-aware summaries, where R{\footnotesize OUGE}-1/2/L increases by +5.05/+1.35/+4.33 points on \textit{CNN/DailyMail} and +3.96/+4.36/+4.77 points on \textit{BBC XSum}, demonstrating that GPT-3 successfully focuses on more core elements through SumCoT and further fits the element-aware writing pattern.

Finally, we also conduct human studies to compare summaries of GPT-3 w/o and w/ SumCoT. Results (as shown in Table \ref{tab:human_study_cot}) indicate that the SumCoT technique further improves the performance of the standard zero-shot paradigm in all dimensions, particularly coherence and relevance.

\subsection{Better Understanding SumCoT}\label{sec:analysis}

\begin{figure*}[t]
  \centering
  \includegraphics[width=1.00\linewidth]{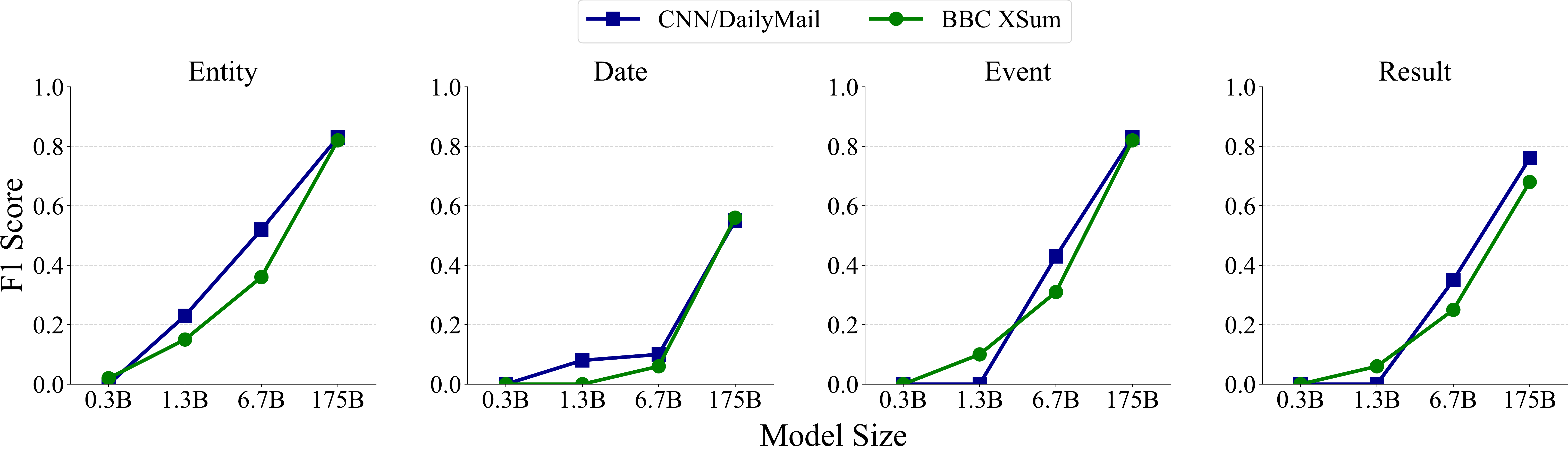}  
  \vspace{-0.25in}
  \caption{Performance of element extraction for all four core elements with various GPT-3 versions. See Appendix \ref{sec:ablation} for more model details.}
  \label{img:modelsize}
\vspace{-3mm}
\end{figure*}

\paragraph{How does SumCoT affect summary writing?}  First, we explore the extent to which SumCoT affects the final summary generation.
We compute the coverage, the fraction of extracted elements in Stage 1 actually appearing in the final summary generated in Stage 2. 
Table \ref{tab:sumcot_affect} shows the results (see Appendix \ref{sec:gpt3_vs_gpt3sumcot} for examples), and we observe that final summaries are extremely faithful to the extracted elements, particularly on \textit{CNN/DailyMail}. On \textit{BBC XSum}, the coverages of each element are relatively lower due to the one-sentence style of \textit{BBC XSum}, resulting in further condensation of the extracted elements. 
In addition, the coverage of \textit{\textbf{Date}} is significantly low, probably due to the errors of extraction. This will be verified in the next part.

\begin{table}[t]
\centering
\footnotesize
  \renewcommand\arraystretch{1}
  \setlength{\tabcolsep}{1.2mm}
  {
  \resizebox{\columnwidth}{!}{
\begin{tabular}{cccc|cccc}
\toprule

\multicolumn{4}{c|}{\textbf{\textit{CNN/DailyMail}}} & \multicolumn{4}{c}{\textbf{\textit{BBC XSum}}}\\
\midrule

Entity & Date & Event & Result & Entity & Date & Event & Result\\

0.89 & 0.55 & 0.93 & 0.95 & 0.80 & 0.48 & 0.87 & 0.66\\

\bottomrule                     
        
\end{tabular}}
    \caption{Coverage, the fraction of extracted elements actually appearing in the final summary on two datasets.}
    \label{tab:sumcot_affect}%
}

\end{table}

\begin{table}[t]
\centering
\footnotesize
  \renewcommand\arraystretch{1}
  \setlength{\tabcolsep}{2.0mm}{
  \resizebox{0.95\columnwidth}{!}{
\begin{tabular}{l|ccc|ccc}
\toprule

\multirow{2}{*}{\makecell[l]{\textbf{Core}\\\textbf{Element}}} 
& \multicolumn{3}{c|}{\textbf{\textit{CNN/DaliyMail}}} 
& \multicolumn{3}{c}{\textbf{\textit{BBC XSum}}}\\
\cmidrule(r){2-7} 

& \textit{P} & \textit{R} & \textbf{\textit{F1}} & \textit{P} & \textit{R} & \textbf{\textit{F1}}\\
          
\midrule

Entity & 0.77 & 0.89 & \textbf{0.83} & 0.71 & 0.98 & \textbf{0.82} \\
Date & 0.46 & 0.68 & \textbf{0.55} & 0.43 & 0.79 & \textbf{0.56}\\
Event & 0.84 & 0.82 & \textbf{0.83} & 0.75 & 0.90 & \textbf{0.82}\\
Result & 0.74 & 0.79 & \textbf{0.76} & 0.66 & 0.71 & \textbf{0.68}\\

\bottomrule

\end{tabular}}
    \caption{The $\mathrm{Precision}$ (\textit{P}), $\mathrm{Recall}$ (\textit{R}), and $F_1$ for extraction of each element.}
    \label{tab:extraction}%
}

\end{table}

\paragraph{Is the element extraction accurate and comprehensive?}  Table \ref{tab:sumcot_affect} demonstrates a strong correlation between element extraction and summary generation, so we need to examine the quality of element extraction.\footnote{It is noted that if there are no obvious markers (e.g. "\textit{The entities are ...}"), the extraction is not considered valid.}
We compute the $\mathrm{Precision}$, $\mathrm{Recall}$ and $F_1$ introduced in §\ref{sec:elementaware}. Results (Table \ref{tab:extraction}) show that extraction achieves an outperforming result except for \textit{\textbf{Date}}, and $\mathrm{Precision}$ are usually lower than $\mathrm{Recall}$. See Appendix \ref{sec:errorcase} for error cases, where we conclude: (i) Date hallucination is particularly evident for extracting non-existent dates; (ii) Element redundancy often occurs.

\paragraph{Does the model size limit SumCoT?}
We compare the performance of GPT-3 with different versions of element extraction. We compute the $F_1$ score (shown in Figure \ref{img:modelsize}) for all the elements.
We find that when the model size is small, element extraction is almost invalid. As the model size increases, GPT-3 can extract one by one for all types of elements, but the extraction itself has many errors or redundancies. Only when the model size is the largest, the element extraction is human-approved
(See Appendix \ref{sec:ablation} for examples).
This indicates that the SumCoT technique is also an emergent ability of model scale \cite{wei2022emergent}, and is effective only when the model size is larger.

\section{Related Work and Discussion}

\subsection{Summarization: Dataset and Evaluation}  In the data-driven deep learning era, large-scale corpus crawled from websites for summarization is rich, especially the news domain. 
They can be divided into the single-document setting \cite{harman2004effects,NYT2008,napoles2012annotated,CNNDM,XSUM,wikihow,grusky2018newsroom} and the multi-document setting \cite{TAC2011,li2017reader,multinews} according to the source numbers of document clusters.
However, some studies pointed out various noises within them, such as poor coherence, information redundancy, and factual hallucination \cite{kryscinski2019neural,maynez2020faithfulness,fabbri2021summeval}. Several other studies also corroborated this with human assessments \cite{stiennon2020learning,fabbri2021summeval}.

Summarization systems are first purely trained \cite{vinyals2015pointer,vaswani2017attention,liu2022brio,chen2022unisumm} or fine-tuned  \cite{zhang2019hibert,liu2019fine,zhang2020pegasus,2020t5,wang2022noise,mao2022explicitly} with standard datasets, and then evaluated.
The most mainstream automatic evaluation metrics for summarization are reference-based methods, i.e., directly comparing the similarity of generated and dataset-specific summaries. They can be split into lexical overlap methods \cite{papineni2002bleu,lin2004rouge,banerjee2005meteor} and semantic similarity methods \cite{ng2015better,zhang2019bertscore,zhao2019moverscore,sellam2020bleurt,rei2020comet}. Such evaluation is essentially a test of the fit degree to standard datasets.
In recent years, the advanced zero-shot paradigm of LLMs makes text generation free of standard datasets \cite{gpt3,chowdhery2022palm,thoppilan2022lamda} but rely on massive pre-trained data, many researchers tend to revisit the quality assessment of summaries generated by LLMs \cite{liu2022revisiting,zhang2023benchmarking}. However, some studies demonstrate that automatic evaluation results do not align with human preference in summarization tasks \cite{goyal2022news}, similar counter-intuitive observations may pose new challenges for the evaluation in the era of LLMs.

\subsection{Chain-of-Thought Prompting for LLMs}
Recently, intriguing chain-of-thought techniques have greatly improved both the reasoning performance and interpretability of LLMs by decomposing multi-step problems into intermediate steps \cite{nye2022show,wei2022chain,kojima2022large,zhang2022automatic,wang2022self,zhang2023multimodal,shi2022language,zhou2022least}.
However, no prior work has studied CoT in the scenario of automatic summarization. To the best of our knowledge, we are the first to study chain-of-thought prompting for summarization, eliciting LLMs to leverage more fine-grained elements from source documents to generate effective summaries.

\section{Conclusion}

In this work, we construct expert-writing element-aware summary test sets for \textit{CNN/DailyMail} and \textit{
BBC XSum}, they are specifically designed to assess the generic summarization capabilities of diverse, powerful language models more thoroughly.
Upon the fine-grained test sets, we preliminarily conduct experiments on zero-shot LLMs and fine-tuned PLMs, demonstrating the surprising zero-shot summary writing ability of LLMs. 
Further, we propose a CoT-based method, which elicits LLMs to focus on core news elements and generate summaries step by step.
In the future, we hope that our work will inspire further research into harnessing LLMs' potential to mimic human writing processes across various open-ended generative tasks.

\section*{Limitations}

In terms of the test sets, due to time, labor, and financial limitations, we are unable to construct large-scale test sets of the same size as the original, so the domain balance in the test sets is not fully considered, but the uniformity of writing style might have slightly alleviated this issue.
In terms of the method, we empirically explore the possibility of chain-of-thought application in text generation. However, due to the stronger openness of generative tasks compared to pure reasoning tasks, generated summaries might be more sensitive to the form of chain-of-thought, which is a key point worth further optimization.

\section*{Ethics Statement}

We use publicly available source documents from existing general datasets for annotations, so the ethics issues of the source texts are non-existent.
For the generated contents with LLMs, e.g. GPT-3, prior work \cite{gpt3,chan2022gpt} has elaborated on their inevitable potential toxicity, such as issues of bias and fairness. Moreover, this is the first work to apply the chain-of-thought technique to open-end generation tasks, so we completely keep the prompts neutral and task-specific to avoid toxic language generation, and there were no toxic texts that appeared in our experiments.

\section*{Acknowledgements}

Yiming and Rui are with MT-Lab, Department of Computer Science and Engineering,
School of Electronic Information and Electrical Engineering, and also with the MoE Key Lab of Artificial Intelligence, AI Institute, Shanghai Jiao Tong
University, Shanghai 200204, China. Rui is supported by the General Program of National Natural Science Foundation of China (6217020129), Shanghai
Pujiang Program (21PJ1406800), Shanghai
Municipal Science and Technology Major Project
(2021SHZDZX0102), Beijing Academy of Artificial Intelligence (BAAI) (No. 4), CCF-Baidu Open Fund (F2022018), and the Alibaba-AIR Program (22088682). We also thank the computational resource from the SJTU student innovation center.

\bibliography{anthology,output}

\begin{thebibliography}{62}
\expandafter\ifx\csname natexlab\endcsname\relax\def\natexlab#1{#1}\fi

\bibitem[{Banerjee and Lavie(2005)}]{banerjee2005meteor}
Satanjeev Banerjee and Alon Lavie. 2005.
\newblock \href {https://aclanthology.org/W05-0909} {{METEOR}: An automatic
  metric for {MT} evaluation with improved correlation with human judgments}.
\newblock In \emph{Proceedings of the {ACL} Workshop on Intrinsic and Extrinsic
  Evaluation Measures for Machine Translation and/or Summarization}, pages
  65--72, Ann Arbor, Michigan. Association for Computational Linguistics.

\bibitem[{Brown et~al.(2020)Brown, Mann, Ryder, Subbiah, Kaplan, Dhariwal,
  Neelakantan, Shyam, Sastry, Askell, Agarwal, Herbert{-}Voss, Krueger,
  Henighan, Child, Ramesh, Ziegler, Wu, Winter, Hesse, Chen, Sigler, Litwin,
  Gray, Chess, Clark, Berner, McCandlish, Radford, Sutskever, and
  Amodei}]{gpt3}
Tom~B. Brown, Benjamin Mann, Nick Ryder, Melanie Subbiah, Jared Kaplan,
  Prafulla Dhariwal, Arvind Neelakantan, Pranav Shyam, Girish Sastry, Amanda
  Askell, Sandhini Agarwal, Ariel Herbert{-}Voss, Gretchen Krueger, Tom
  Henighan, Rewon Child, Aditya Ramesh, Daniel~M. Ziegler, Jeffrey Wu, Clemens
  Winter, Christopher Hesse, Mark Chen, Eric Sigler, Mateusz Litwin, Scott
  Gray, Benjamin Chess, Jack Clark, Christopher Berner, Sam McCandlish, Alec
  Radford, Ilya Sutskever, and Dario Amodei. 2020.
\newblock \href
  {https://proceedings.neurips.cc/paper/2020/hash/1457c0d6bfcb4967418bfb8ac142f64a-Abstract.html}
  {Language models are few-shot learners}.
\newblock In \emph{Advances in Neural Information Processing Systems 33: Annual
  Conference on Neural Information Processing Systems 2020, NeurIPS 2020,
  December 6-12, 2020, virtual}.

\bibitem[{Chan(2022)}]{chan2022gpt}
Anastasia Chan. 2022.
\newblock \href {https://link.springer.com/article/10.1007/s43681-022-00148-6}
  {Gpt-3 and instructgpt: technological dystopianism, utopianism, and
  “contextual” perspectives in ai ethics and industry}.
\newblock \emph{AI and Ethics}, pages 1--12.

\bibitem[{Chen et~al.(2022)Chen, Liu, Xu, Yang, Zhu, Zeng, and
  Zhang}]{chen2022unisumm}
Yulong Chen, Yang Liu, Ruochen Xu, Ziyi Yang, Chenguang Zhu, Michael Zeng, and
  Yue Zhang. 2022.
\newblock \href {https://arxiv.org/abs/2211.09783} {Unisumm: Unified few-shot
  summarization with multi-task pre-training and prefix-tuning}.
\newblock \emph{ArXiv preprint}, abs/2211.09783.

\bibitem[{Chowdhery et~al.(2022)Chowdhery, Narang, Devlin, Bosma, Mishra,
  Roberts, Barham, Chung, Sutton, Gehrmann et~al.}]{chowdhery2022palm}
Aakanksha Chowdhery, Sharan Narang, Jacob Devlin, Maarten Bosma, Gaurav Mishra,
  Adam Roberts, Paul Barham, Hyung~Won Chung, Charles Sutton, Sebastian
  Gehrmann, et~al. 2022.
\newblock \href {https://arxiv.org/abs/2204.02311} {Palm: Scaling language
  modeling with pathways}.
\newblock \emph{ArXiv preprint}, abs/2204.02311.

\bibitem[{Devlin et~al.(2019)Devlin, Chang, Lee, and
  Toutanova}]{devlin2019bert}
Jacob Devlin, Ming-Wei Chang, Kenton Lee, and Kristina Toutanova. 2019.
\newblock \href {https://doi.org/10.18653/v1/N19-1423} {{BERT}: Pre-training of
  deep bidirectional transformers for language understanding}.
\newblock In \emph{Proceedings of the 2019 Conference of the North {A}merican
  Chapter of the Association for Computational Linguistics: Human Language
  Technologies, Volume 1 (Long and Short Papers)}, pages 4171--4186,
  Minneapolis, Minnesota. Association for Computational Linguistics.

\bibitem[{Fabbri et~al.(2019)Fabbri, Li, She, Li, and Radev}]{multinews}
Alexander Fabbri, Irene Li, Tianwei She, Suyi Li, and Dragomir Radev. 2019.
\newblock \href {https://doi.org/10.18653/v1/P19-1102} {Multi-news: A
  large-scale multi-document summarization dataset and abstractive hierarchical
  model}.
\newblock In \emph{Proceedings of the 57th Annual Meeting of the Association
  for Computational Linguistics}, pages 1074--1084, Florence, Italy.
  Association for Computational Linguistics.

\bibitem[{Fabbri et~al.(2021)Fabbri, Kry{\'s}ci{\'n}ski, McCann, Xiong, Socher,
  and Radev}]{fabbri2021summeval}
Alexander~R. Fabbri, Wojciech Kry{\'s}ci{\'n}ski, Bryan McCann, Caiming Xiong,
  Richard Socher, and Dragomir Radev. 2021.
\newblock \href {https://doi.org/10.1162/tacl_a_00373} {{S}umm{E}val:
  Re-evaluating summarization evaluation}.
\newblock \emph{Transactions of the Association for Computational Linguistics},
  9:391--409.

\bibitem[{Gehrmann et~al.(2018)Gehrmann, Deng, and Rush}]{gehrmann2018bottom}
Sebastian Gehrmann, Yuntian Deng, and Alexander Rush. 2018.
\newblock \href {https://doi.org/10.18653/v1/D18-1443} {Bottom-up abstractive
  summarization}.
\newblock In \emph{Proceedings of the 2018 Conference on Empirical Methods in
  Natural Language Processing}, pages 4098--4109, Brussels, Belgium.
  Association for Computational Linguistics.

\bibitem[{Goyal et~al.(2022)Goyal, Li, and Durrett}]{goyal2022news}
Tanya Goyal, Junyi~Jessy Li, and Greg Durrett. 2022.
\newblock \href {https://arxiv.org/abs/2209.12356} {News summarization and
  evaluation in the era of gpt-3}.
\newblock \emph{ArXiv preprint}, abs/2209.12356.

\bibitem[{Grusky et~al.(2018)Grusky, Naaman, and Artzi}]{grusky2018newsroom}
Max Grusky, Mor Naaman, and Yoav Artzi. 2018.
\newblock \href {https://doi.org/10.18653/v1/N18-1065} {{N}ewsroom: A dataset
  of 1.3 million summaries with diverse extractive strategies}.
\newblock In \emph{Proceedings of the 2018 Conference of the North {A}merican
  Chapter of the Association for Computational Linguistics: Human Language
  Technologies, Volume 1 (Long Papers)}, pages 708--719, New Orleans,
  Louisiana. Association for Computational Linguistics.

\bibitem[{Harman and Over(2004)}]{harman2004effects}
Donna Harman and Paul Over. 2004.
\newblock \href {https://aclanthology.org/W04-1003} {The effects of human
  variation in {DUC} summarization evaluation}.
\newblock In \emph{Text Summarization Branches Out}, pages 10--17, Barcelona,
  Spain. Association for Computational Linguistics.

\bibitem[{He et~al.(2020)He, Kry{\'s}ci{\'n}ski, McCann, Rajani, and
  Xiong}]{he2020ctrlsum}
Junxian He, Wojciech Kry{\'s}ci{\'n}ski, Bryan McCann, Nazneen Rajani, and
  Caiming Xiong. 2020.
\newblock \href {https://arxiv.org/abs/2012.04281} {Ctrlsum: Towards generic
  controllable text summarization}.
\newblock \emph{ArXiv preprint}, abs/2012.04281.

\bibitem[{Hermann et~al.(2015)Hermann, Kocisk{\'{y}}, Grefenstette, Espeholt,
  Kay, Suleyman, and Blunsom}]{hermann2015teaching}
Karl~Moritz Hermann, Tom{\'{a}}s Kocisk{\'{y}}, Edward Grefenstette, Lasse
  Espeholt, Will Kay, Mustafa Suleyman, and Phil Blunsom. 2015.
\newblock \href
  {https://proceedings.neurips.cc/paper/2015/hash/afdec7005cc9f14302cd0474fd0f3c96-Abstract.html}
  {Teaching machines to read and comprehend}.
\newblock In \emph{Advances in Neural Information Processing Systems 28: Annual
  Conference on Neural Information Processing Systems 2015, December 7-12,
  2015, Montreal, Quebec, Canada}, pages 1693--1701.

\bibitem[{Kojima et~al.(2022)Kojima, Gu, Reid, Matsuo, and
  Iwasawa}]{kojima2022large}
Takeshi Kojima, Shixiang~Shane Gu, Machel Reid, Yutaka Matsuo, and Yusuke
  Iwasawa. 2022.
\newblock \href {https://arxiv.org/abs/2205.11916} {Large language models are
  zero-shot reasoners}.
\newblock \emph{ArXiv preprint}, abs/2205.11916.

\bibitem[{Koupaee and Wang(2018)}]{wikihow}
Mahnaz Koupaee and William~Yang Wang. 2018.
\newblock \href {https://arxiv.org/abs/1810.09305} {Wikihow: A large scale text
  summarization dataset}.
\newblock \emph{ArXiv preprint}, abs/1810.09305.

\bibitem[{Kryscinski et~al.(2019)Kryscinski, Keskar, McCann, Xiong, and
  Socher}]{kryscinski2019neural}
Wojciech Kryscinski, Nitish~Shirish Keskar, Bryan McCann, Caiming Xiong, and
  Richard Socher. 2019.
\newblock \href {https://doi.org/10.18653/v1/D19-1051} {Neural text
  summarization: A critical evaluation}.
\newblock In \emph{Proceedings of the 2019 Conference on Empirical Methods in
  Natural Language Processing and the 9th International Joint Conference on
  Natural Language Processing (EMNLP-IJCNLP)}, pages 540--551, Hong Kong,
  China. Association for Computational Linguistics.

\bibitem[{Lasswell(1948)}]{lasswell1948structure}
Harold~D Lasswell. 1948.
\newblock \href {https://www.scinapse.io/papers/2290526371} {The structure and
  function of communication in society}.
\newblock \emph{The communication of ideas}, 37(1):136--139.

\bibitem[{Lewis et~al.(2020)Lewis, Liu, Goyal, Ghazvininejad, Mohamed, Levy,
  Stoyanov, and Zettlemoyer}]{lewis2020bart}
Mike Lewis, Yinhan Liu, Naman Goyal, Marjan Ghazvininejad, Abdelrahman Mohamed,
  Omer Levy, Veselin Stoyanov, and Luke Zettlemoyer. 2020.
\newblock \href {https://doi.org/10.18653/v1/2020.acl-main.703} {{BART}:
  Denoising sequence-to-sequence pre-training for natural language generation,
  translation, and comprehension}.
\newblock In \emph{Proceedings of the 58th Annual Meeting of the Association
  for Computational Linguistics}, pages 7871--7880, Online. Association for
  Computational Linguistics.

\bibitem[{Li et~al.(2017)Li, Bing, and Lam}]{li2017reader}
Piji Li, Lidong Bing, and Wai Lam. 2017.
\newblock \href {https://doi.org/10.18653/v1/W17-4512} {Reader-aware
  multi-document summarization: An enhanced model and the first dataset}.
\newblock In \emph{Proceedings of the Workshop on New Frontiers in
  Summarization}, pages 91--99, Copenhagen, Denmark. Association for
  Computational Linguistics.

\bibitem[{Likert(1932)}]{likert1932technique}
Rensis Likert. 1932.
\newblock \href {https://psycnet.apa.org/record/1933-01885-001} {A technique
  for the measurement of attitudes.}
\newblock \emph{Archives of psychology}.

\bibitem[{Lin(2004)}]{lin2004rouge}
Chin-Yew Lin. 2004.
\newblock \href {https://aclanthology.org/W04-1013} {{ROUGE}: A package for
  automatic evaluation of summaries}.
\newblock In \emph{Text Summarization Branches Out}, pages 74--81, Barcelona,
  Spain. Association for Computational Linguistics.

\bibitem[{Liu(2019)}]{liu2019fine}
Yang Liu. 2019.
\newblock \href {https://arxiv.org/abs/1903.10318} {Fine-tune bert for
  extractive summarization}.
\newblock \emph{ArXiv preprint}, abs/1903.10318.

\bibitem[{Liu et~al.(2019)Liu, Ott, Goyal, Du, Joshi, Chen, Levy, Lewis,
  Zettlemoyer, and Stoyanov}]{liu2019roberta}
Yinhan Liu, Myle Ott, Naman Goyal, Jingfei Du, Mandar Joshi, Danqi Chen, Omer
  Levy, Mike Lewis, Luke Zettlemoyer, and Veselin Stoyanov. 2019.
\newblock \href {https://arxiv.org/abs/1907.11692} {Roberta: A robustly
  optimized bert pretraining approach}.
\newblock \emph{ArXiv preprint}, abs/1907.11692.

\bibitem[{Liu et~al.(2022{\natexlab{a}})Liu, Fabbri, Liu, Zhao, Nan, Han, Han,
  Joty, Wu, Xiong et~al.}]{liu2022revisiting}
Yixin Liu, Alexander~R Fabbri, Pengfei Liu, Yilun Zhao, Linyong Nan, Ruilin
  Han, Simeng Han, Shafiq Joty, Chien-Sheng Wu, Caiming Xiong, et~al.
  2022{\natexlab{a}}.
\newblock \href {https://arxiv.org/abs/2212.07981} {Revisiting the gold
  standard: Grounding summarization evaluation with robust human evaluation}.
\newblock \emph{arXiv e-prints}, pages arXiv--2212.

\bibitem[{Liu et~al.(2022{\natexlab{b}})Liu, Liu, Radev, and
  Neubig}]{liu2022brio}
Yixin Liu, Pengfei Liu, Dragomir Radev, and Graham Neubig. 2022{\natexlab{b}}.
\newblock \href {https://doi.org/10.18653/v1/2022.acl-long.207} {{BRIO}:
  Bringing order to abstractive summarization}.
\newblock In \emph{Proceedings of the 60th Annual Meeting of the Association
  for Computational Linguistics (Volume 1: Long Papers)}, pages 2890--2903,
  Dublin, Ireland. Association for Computational Linguistics.

\bibitem[{Mao et~al.(2022)Mao, Li, Wang, Li, Hao, Wang, and
  Wang}]{mao2022explicitly}
Qianren Mao, Jianxin Li, JiaZheng Wang, Xi~Li, Peng Hao, Lihong Wang, and Zheng
  Wang. 2022.
\newblock \href {https://ieeexplore.ieee.org/abstract/document/9746383}
  {Explicitly modeling importance and coherence for timeline summarization}.
\newblock In \emph{ICASSP 2022-2022 IEEE International Conference on Acoustics,
  Speech and Signal Processing (ICASSP)}, pages 8062--8066. IEEE.

\bibitem[{Maynez et~al.(2020)Maynez, Narayan, Bohnet, and
  McDonald}]{maynez2020faithfulness}
Joshua Maynez, Shashi Narayan, Bernd Bohnet, and Ryan McDonald. 2020.
\newblock \href {https://doi.org/10.18653/v1/2020.acl-main.173} {On
  faithfulness and factuality in abstractive summarization}.
\newblock In \emph{Proceedings of the 58th Annual Meeting of the Association
  for Computational Linguistics}, pages 1906--1919, Online. Association for
  Computational Linguistics.

\bibitem[{Nallapati et~al.(2016)Nallapati, Zhou, dos Santos, Gu̇l{\c{c}}ehre,
  and Xiang}]{CNNDM}
Ramesh Nallapati, Bowen Zhou, Cicero dos Santos, {\c{C}}a{\u{g}}lar
  Gu̇l{\c{c}}ehre, and Bing Xiang. 2016.
\newblock \href {https://doi.org/10.18653/v1/K16-1028} {Abstractive text
  summarization using sequence-to-sequence {RNN}s and beyond}.
\newblock In \emph{Proceedings of the 20th {SIGNLL} Conference on Computational
  Natural Language Learning}, pages 280--290, Berlin, Germany. Association for
  Computational Linguistics.

\bibitem[{Napoles et~al.(2012)Napoles, Gormley, and
  Van~Durme}]{napoles2012annotated}
Courtney Napoles, Matthew Gormley, and Benjamin Van~Durme. 2012.
\newblock \href {https://aclanthology.org/W12-3018} {Annotated {G}igaword}.
\newblock In \emph{Proceedings of the Joint Workshop on Automatic Knowledge
  Base Construction and Web-scale Knowledge Extraction ({AKBC}-{WEKEX})}, pages
  95--100, Montr{\'e}al, Canada. Association for Computational Linguistics.

\bibitem[{Narayan et~al.(2018)Narayan, Cohen, and Lapata}]{XSUM}
Shashi Narayan, Shay~B. Cohen, and Mirella Lapata. 2018.
\newblock \href {https://doi.org/10.18653/v1/D18-1206} {Don{'}t give me the
  details, just the summary! topic-aware convolutional neural networks for
  extreme summarization}.
\newblock In \emph{Proceedings of the 2018 Conference on Empirical Methods in
  Natural Language Processing}, pages 1797--1807, Brussels, Belgium.
  Association for Computational Linguistics.

\bibitem[{Narayan et~al.(2022)Narayan, Sim{\~o}es, Zhao, Maynez, Das, Collins,
  and Lapata}]{narayan2022well}
Shashi Narayan, Gon{\c{c}}alo Sim{\~o}es, Yao Zhao, Joshua Maynez, Dipanjan
  Das, Michael Collins, and Mirella Lapata. 2022.
\newblock \href {https://doi.org/10.18653/v1/2022.acl-long.94} {A well-composed
  text is half done! composition sampling for diverse conditional generation}.
\newblock In \emph{Proceedings of the 60th Annual Meeting of the Association
  for Computational Linguistics (Volume 1: Long Papers)}, pages 1319--1339,
  Dublin, Ireland. Association for Computational Linguistics.

\bibitem[{Narayan et~al.(2021)Narayan, Zhao, Maynez, Sim{\~o}es, Nikolaev, and
  McDonald}]{narayan2021planning}
Shashi Narayan, Yao Zhao, Joshua Maynez, Gon{\c{c}}alo Sim{\~o}es, Vitaly
  Nikolaev, and Ryan McDonald. 2021.
\newblock \href {https://doi.org/10.1162/tacl_a_00438} {Planning with learned
  entity prompts for abstractive summarization}.
\newblock \emph{Transactions of the Association for Computational Linguistics},
  9:1475--1492.

\bibitem[{Ng and Abrecht(2015)}]{ng2015better}
Jun-Ping Ng and Viktoria Abrecht. 2015.
\newblock \href {https://doi.org/10.18653/v1/D15-1222} {Better summarization
  evaluation with word embeddings for {ROUGE}}.
\newblock In \emph{Proceedings of the 2015 Conference on Empirical Methods in
  Natural Language Processing}, pages 1925--1930, Lisbon, Portugal. Association
  for Computational Linguistics.

\bibitem[{Nye et~al.(2022)Nye, Andreassen, Gur-Ari, Michalewski, Austin,
  Bieber, Dohan, Lewkowycz, Bosma, Luan et~al.}]{nye2022show}
Maxwell Nye, Anders~Johan Andreassen, Guy Gur-Ari, Henryk Michalewski, Jacob
  Austin, David Bieber, David Dohan, Aitor Lewkowycz, Maarten Bosma, David
  Luan, et~al. 2022.
\newblock \href {https://openreview.net/forum?id=HBlx2idbkbq} {Show your work:
  Scratchpads for intermediate computation with language models}.
\newblock In \emph{Deep Learning for Code Workshop}.

\bibitem[{Ouyang et~al.(2022)Ouyang, Wu, Jiang, Almeida, Wainwright, Mishkin,
  Zhang, Agarwal, Slama, Ray et~al.}]{ouyang2022training}
Long Ouyang, Jeff Wu, Xu~Jiang, Diogo Almeida, Carroll~L Wainwright, Pamela
  Mishkin, Chong Zhang, Sandhini Agarwal, Katarina Slama, Alex Ray, et~al.
  2022.
\newblock \href {https://arxiv.org/abs/2203.02155} {Training language models to
  follow instructions with human feedback}.
\newblock \emph{ArXiv preprint}, abs/2203.02155.

\bibitem[{Owczarzak and Dang(2011)}]{TAC2011}
Karolina Owczarzak and Hoa~Trang Dang. 2011.
\newblock Overview of the tac 2011 summarization track: Guided task and aesop
  task.
\newblock In \emph{Proceedings of the Text Analysis Conference (TAC 2011),
  Gaithersburg, Maryland, USA, November}.

\bibitem[{Papineni et~al.(2002)Papineni, Roukos, Ward, and
  Zhu}]{papineni2002bleu}
Kishore Papineni, Salim Roukos, Todd Ward, and Wei-Jing Zhu. 2002.
\newblock \href {https://doi.org/10.3115/1073083.1073135} {{B}leu: a method for
  automatic evaluation of machine translation}.
\newblock In \emph{Proceedings of the 40th Annual Meeting of the Association
  for Computational Linguistics}, pages 311--318, Philadelphia, Pennsylvania,
  USA. Association for Computational Linguistics.

\bibitem[{P{\"{o}}~ttker(2003)}]{po2003news}
Horst P{\"{o}}~ttker. 2003.
\newblock \href
  {https://www.tandfonline.com/doi/abs/10.1080/1461670032000136596} {News and
  its communicative quality: the inverted pyramid—when and why did it
  appear?}
\newblock \emph{Journalism Studies}, 4(4):501--511.

\bibitem[{Raffel et~al.(2020)Raffel, Shazeer, Roberts, Lee, Narang, Matena,
  Zhou, Li, and Liu}]{2020t5}
Colin Raffel, Noam Shazeer, Adam Roberts, Katherine Lee, Sharan Narang, Michael
  Matena, Yanqi Zhou, Wei Li, and Peter~J. Liu. 2020.
\newblock \href {http://jmlr.org/papers/v21/20-074.html} {Exploring the limits
  of transfer learning with a unified text-to-text transformer}.
\newblock \emph{Journal of Machine Learning Research}, 21(140):1--67.

\bibitem[{Rei et~al.(2020)Rei, Stewart, Farinha, and Lavie}]{rei2020comet}
Ricardo Rei, Craig Stewart, Ana~C Farinha, and Alon Lavie. 2020.
\newblock \href {https://doi.org/10.18653/v1/2020.emnlp-main.213} {{COMET}: A
  neural framework for {MT} evaluation}.
\newblock In \emph{Proceedings of the 2020 Conference on Empirical Methods in
  Natural Language Processing (EMNLP)}, pages 2685--2702, Online. Association
  for Computational Linguistics.

\bibitem[{Sandhaus(2008)}]{NYT2008}
Evan Sandhaus. 2008.
\newblock \href {https://catalog.ldc.upenn.edu/LDC2008T19} {The new york times
  annotated corpus}.
\newblock In \emph{Philadelphia: Linguistic Data Consortium, 2008.}

\bibitem[{Sanh et~al.(2022)Sanh, Webson, Raffel, Bach, Sutawika, Alyafeai,
  Chaffin, Stiegler, Scao, Raja et~al.}]{sanh2022multitask}
Victor Sanh, Albert Webson, Colin Raffel, Stephen Bach, Lintang Sutawika, Zaid
  Alyafeai, Antoine Chaffin, Arnaud Stiegler, Teven Scao, Arun Raja, et~al.
  2022.
\newblock \href {https://arxiv.org/abs/2110.08207} {Multitask prompted training
  enables zero-shot task generalization}.
\newblock In \emph{International Conference on Learning Representations}.

\bibitem[{Sellam et~al.(2020)Sellam, Das, and Parikh}]{sellam2020bleurt}
Thibault Sellam, Dipanjan Das, and Ankur Parikh. 2020.
\newblock \href {https://doi.org/10.18653/v1/2020.acl-main.704} {{BLEURT}:
  Learning robust metrics for text generation}.
\newblock In \emph{Proceedings of the 58th Annual Meeting of the Association
  for Computational Linguistics}, pages 7881--7892, Online. Association for
  Computational Linguistics.

\bibitem[{Shi et~al.(2022)Shi, Suzgun, Freitag, Wang, Srivats, Vosoughi, Chung,
  Tay, Ruder, Zhou et~al.}]{shi2022language}
Freda Shi, Mirac Suzgun, Markus Freitag, Xuezhi Wang, Suraj Srivats, Soroush
  Vosoughi, Hyung~Won Chung, Yi~Tay, Sebastian Ruder, Denny Zhou, et~al. 2022.
\newblock \href {https://arxiv.org/abs/2210.03057} {Language models are
  multilingual chain-of-thought reasoners}.
\newblock \emph{ArXiv preprint}, abs/2210.03057.

\bibitem[{Stiennon et~al.(2020)Stiennon, Ouyang, Wu, Ziegler, Lowe, Voss,
  Radford, Amodei, and Christiano}]{stiennon2020learning}
Nisan Stiennon, Long Ouyang, Jeffrey Wu, Daniel~M. Ziegler, Ryan Lowe, Chelsea
  Voss, Alec Radford, Dario Amodei, and Paul~F. Christiano. 2020.
\newblock \href
  {https://proceedings.neurips.cc/paper/2020/hash/1f89885d556929e98d3ef9b86448f951-Abstract.html}
  {Learning to summarize with human feedback}.
\newblock In \emph{Advances in Neural Information Processing Systems 33: Annual
  Conference on Neural Information Processing Systems 2020, NeurIPS 2020,
  December 6-12, 2020, virtual}.

\bibitem[{Sutskever et~al.(2014)Sutskever, Vinyals, and
  Le}]{sutskever2014sequence}
Ilya Sutskever, Oriol Vinyals, and Quoc~V. Le. 2014.
\newblock \href
  {https://proceedings.neurips.cc/paper/2014/hash/a14ac55a4f27472c5d894ec1c3c743d2-Abstract.html}
  {Sequence to sequence learning with neural networks}.
\newblock In \emph{Advances in Neural Information Processing Systems 27: Annual
  Conference on Neural Information Processing Systems 2014, December 8-13 2014,
  Montreal, Quebec, Canada}, pages 3104--3112.

\bibitem[{Thoppilan et~al.(2022)Thoppilan, De~Freitas, Hall, Shazeer,
  Kulshreshtha, Cheng, Jin, Bos, Baker, Du et~al.}]{thoppilan2022lamda}
Romal Thoppilan, Daniel De~Freitas, Jamie Hall, Noam Shazeer, Apoorv
  Kulshreshtha, Heng-Tze Cheng, Alicia Jin, Taylor Bos, Leslie Baker, Yu~Du,
  et~al. 2022.
\newblock \href {https://arxiv.org/abs/2201.08239} {Lamda: Language models for
  dialog applications}.
\newblock \emph{ArXiv preprint}, abs/2201.08239.

\bibitem[{Vaswani et~al.(2017)Vaswani, Shazeer, Parmar, Uszkoreit, Jones,
  Gomez, Kaiser, and Polosukhin}]{vaswani2017attention}
Ashish Vaswani, Noam Shazeer, Niki Parmar, Jakob Uszkoreit, Llion Jones,
  Aidan~N. Gomez, Lukasz Kaiser, and Illia Polosukhin. 2017.
\newblock \href
  {https://proceedings.neurips.cc/paper/2017/hash/3f5ee243547dee91fbd053c1c4a845aa-Abstract.html}
  {Attention is all you need}.
\newblock In \emph{Advances in Neural Information Processing Systems 30: Annual
  Conference on Neural Information Processing Systems 2017, December 4-9, 2017,
  Long Beach, CA, {USA}}, pages 5998--6008.

\bibitem[{Vinyals et~al.(2015)Vinyals, Fortunato, and
  Jaitly}]{vinyals2015pointer}
Oriol Vinyals, Meire Fortunato, and Navdeep Jaitly. 2015.
\newblock \href
  {https://proceedings.neurips.cc/paper/2015/hash/29921001f2f04bd3baee84a12e98098f-Abstract.html}
  {Pointer networks}.
\newblock In \emph{Advances in Neural Information Processing Systems 28: Annual
  Conference on Neural Information Processing Systems 2015, December 7-12,
  2015, Montreal, Quebec, Canada}, pages 2692--2700.

\bibitem[{Wang et~al.(2022{\natexlab{a}})Wang, Wei, Schuurmans, Le, Chi, and
  Zhou}]{wang2022self}
Xuezhi Wang, Jason Wei, Dale Schuurmans, Quoc Le, Ed~Chi, and Denny Zhou.
  2022{\natexlab{a}}.
\newblock \href {https://arxiv.org/abs/2203.11171} {Self-consistency improves
  chain of thought reasoning in language models}.
\newblock \emph{ArXiv preprint}, abs/2203.11171.

\bibitem[{Wang et~al.(2022{\natexlab{b}})Wang, Mao, Liu, Jiang, Zhu, and
  Li}]{wang2022noise}
Yiming Wang, Qianren Mao, Junnan Liu, Weifeng Jiang, Hongdong Zhu, and Jianxin
  Li. 2022{\natexlab{b}}.
\newblock \href {https://aclanthology.org/2022.coling-1.561} {Noise-injected
  consistency training and entropy-constrained pseudo labeling for
  semi-supervised extractive summarization}.
\newblock In \emph{Proceedings of the 29th International Conference on
  Computational Linguistics}, pages 6447--6456, Gyeongju, Republic of Korea.
  International Committee on Computational Linguistics.

\bibitem[{Wei et~al.(2022{\natexlab{a}})Wei, Tay, Bommasani, Raffel, Zoph,
  Borgeaud, Yogatama, Bosma, Zhou, Metzler et~al.}]{wei2022emergent}
Jason Wei, Yi~Tay, Rishi Bommasani, Colin Raffel, Barret Zoph, Sebastian
  Borgeaud, Dani Yogatama, Maarten Bosma, Denny Zhou, Donald Metzler, et~al.
  2022{\natexlab{a}}.
\newblock \href {https://arxiv.org/abs/2206.07682} {Emergent abilities of large
  language models}.
\newblock \emph{ArXiv preprint}, abs/2206.07682.

\bibitem[{Wei et~al.(2022{\natexlab{b}})Wei, Wang, Schuurmans, Bosma, Chi, Le,
  and Zhou}]{wei2022chain}
Jason Wei, Xuezhi Wang, Dale Schuurmans, Maarten Bosma, Ed~Chi, Quoc Le, and
  Denny Zhou. 2022{\natexlab{b}}.
\newblock \href {https://arxiv.org/abs/2201.11903} {Chain of thought prompting
  elicits reasoning in large language models}.
\newblock \emph{ArXiv preprint}, abs/2201.11903.

\bibitem[{Zhang et~al.(2020{\natexlab{a}})Zhang, Zhao, Saleh, and
  Liu}]{zhang2020pegasus}
Jingqing Zhang, Yao Zhao, Mohammad Saleh, and Peter~J. Liu. 2020{\natexlab{a}}.
\newblock \href {http://proceedings.mlr.press/v119/zhang20ae.html} {{PEGASUS:}
  pre-training with extracted gap-sentences for abstractive summarization}.
\newblock In \emph{Proceedings of the 37th International Conference on Machine
  Learning, {ICML} 2020, 13-18 July 2020, Virtual Event}, volume 119 of
  \emph{Proceedings of Machine Learning Research}, pages 11328--11339. {PMLR}.

\bibitem[{Zhang et~al.(2020{\natexlab{b}})Zhang, Kishore, Wu, Weinberger, and
  Artzi}]{zhang2019bertscore}
Tianyi Zhang, Varsha Kishore, Felix Wu, Kilian~Q. Weinberger, and Yoav Artzi.
  2020{\natexlab{b}}.
\newblock \href {https://openreview.net/forum?id=SkeHuCVFDr} {Bertscore:
  Evaluating text generation with {BERT}}.
\newblock In \emph{8th International Conference on Learning Representations,
  {ICLR} 2020, Addis Ababa, Ethiopia, April 26-30, 2020}. OpenReview.net.

\bibitem[{Zhang et~al.(2023{\natexlab{a}})Zhang, Ladhak, Durmus, Liang,
  McKeown, and Hashimoto}]{zhang2023benchmarking}
Tianyi Zhang, Faisal Ladhak, Esin Durmus, Percy Liang, Kathleen McKeown, and
  Tatsunori~B Hashimoto. 2023{\natexlab{a}}.
\newblock \href {https://arxiv.org/abs/2301.13848} {Benchmarking large language
  models for news summarization}.
\newblock \emph{ArXiv preprint}, abs/2301.13848.

\bibitem[{Zhang et~al.(2019)Zhang, Wei, and Zhou}]{zhang2019hibert}
Xingxing Zhang, Furu Wei, and Ming Zhou. 2019.
\newblock \href {https://doi.org/10.18653/v1/P19-1499} {{HIBERT}: Document
  level pre-training of hierarchical bidirectional transformers for document
  summarization}.
\newblock In \emph{Proceedings of the 57th Annual Meeting of the Association
  for Computational Linguistics}, pages 5059--5069, Florence, Italy.
  Association for Computational Linguistics.

\bibitem[{Zhang et~al.(2022)Zhang, Zhang, Li, and Smola}]{zhang2022automatic}
Zhuosheng Zhang, Aston Zhang, Mu~Li, and Alex Smola. 2022.
\newblock \href {https://arxiv.org/abs/2210.03493} {Automatic chain of thought
  prompting in large language models}.
\newblock \emph{ArXiv preprint}, abs/2210.03493.

\bibitem[{Zhang et~al.(2023{\natexlab{b}})Zhang, Zhang, Li, Zhao, Karypis, and
  Smola}]{zhang2023multimodal}
Zhuosheng Zhang, Aston Zhang, Mu~Li, Hai Zhao, George Karypis, and Alex Smola.
  2023{\natexlab{b}}.
\newblock \href {https://arxiv.org/abs/2302.00923} {Multimodal chain-of-thought
  reasoning in language models}.
\newblock \emph{ArXiv preprint, abs/2302.00923}.

\bibitem[{Zhao et~al.(2019)Zhao, Peyrard, Liu, Gao, Meyer, and
  Eger}]{zhao2019moverscore}
Wei Zhao, Maxime Peyrard, Fei Liu, Yang Gao, Christian~M. Meyer, and Steffen
  Eger. 2019.
\newblock \href {https://doi.org/10.18653/v1/D19-1053} {{M}over{S}core: Text
  generation evaluating with contextualized embeddings and earth mover
  distance}.
\newblock In \emph{Proceedings of the 2019 Conference on Empirical Methods in
  Natural Language Processing and the 9th International Joint Conference on
  Natural Language Processing (EMNLP-IJCNLP)}, pages 563--578, Hong Kong,
  China. Association for Computational Linguistics.

\bibitem[{Zhou et~al.(2022)Zhou, Sch{\"a}rli, Hou, Wei, Scales, Wang,
  Schuurmans, Bousquet, Le, and Chi}]{zhou2022least}
Denny Zhou, Nathanael Sch{\"a}rli, Le~Hou, Jason Wei, Nathan Scales, Xuezhi
  Wang, Dale Schuurmans, Olivier Bousquet, Quoc Le, and Ed~Chi. 2022.
\newblock \href {https://arxiv.org/abs/2205.10625} {Least-to-most prompting
  enables complex reasoning in large language models}.
\newblock \emph{ArXiv preprint}, abs/2205.10625.

\end{thebibliography}
\bibliographystyle{acl_natbib}


\appendix

\begin{table*}[t]
\centering
\footnotesize
  \renewcommand\arraystretch{1}
  \setlength{\tabcolsep}{2.5mm}{
  \resizebox{2\columnwidth}{!}{
\begin{tabular}{l|l|l}
\toprule

\multicolumn{1}{c|}{Model} & \multicolumn{1}{c|}{URL} & \multicolumn{1}{c}{License} \\

\midrule

\multirow{4}{*}{B{\scriptsize ART} \cite{lewis2020bart}} &
\url{https://huggingface.co/ainize/bart-base-cnn}
& \multirow{4}{*}{MIT license} \\

& \url{https://huggingface.co/morenolq/bart-base-xsum} & \\
& \url{https://huggingface.co/facebook/bart-large-cnn} & \\
& \url{https://huggingface.co/facebook/bart-large-xsum} & \\
\midrule



\multirow{2}{*}{P{\scriptsize EGASU} \cite{zhang2020pegasus}} & \url{https://huggingface.co/google/pegasus-cnn_dailymail}
 & \multirow{2}{*}{Apache-2.0 license}\\

 & \url{https://huggingface.co/google/pegasus-xsum} & \\

 \midrule

GPT-3 \cite{zhang2020pegasus}  & \url{https://openai.com/api/} & N/A\\

\midrule
\midrule
\multicolumn{1}{c|}{Evaluation Metric} & \multicolumn{1}{c|}{URL} & \multicolumn{1}{c}{License} \\
\midrule

R{\scriptsize OUGE} \cite{lin2004rouge} & \url{https://github.com/pltrdy/rouge} & Apache-2.0 license \\
\midrule
BERTS{\scriptsize CORE} \cite{zhang2019bertscore} & \url{https://github.com/Tiiiger/bert_score} & MIT license \\

\bottomrule

\end{tabular}}
    \caption{The sources and licenses of artifacts and packages we used in this paper (Appendix \ref{sec:maindetail}).}
    \label{tab:artifacts}%
}

\end{table*}

\section{Details of Experimental Setup}\label{sec:details}

\subsection{Main Experiment}\label{sec:maindetail}

Table \ref{tab:artifacts} report the sources and licenses of artifacts and packages we used in this paper.

\subsection{Human Study}
\label{human_study_setup}

We randomly select 50 samples for each dataset and ask three annotators for these tasks following the setting of most human studies. However, considering the unprofessionalism of crowd-sourcing evaluations (Usually hiring workers from Amazon Mechanical
Turk platform with a set hourly salary. Actually, many workers will not work as you expected, their levels vary widely and uncontrollably. \citet{he2020ctrlsum} have encountered such a situation), we privately contact three reliable annotators to conduct the human studies. The first is a Ph.D. candidate in Computer Science, the second is a Master in Film Study, and the last is a graduate in Journalism and Communication. 
Our human studies are conducted in full compliance with the willingness of the invitees and are fully open about the use of the data they annotated.
They have been paid slightly more than the crowd-sourced hourly rate for their work.
We use the same configuration for all human studies in this paper, thanks for their participation!

\section{Abstraction and Faithfulness analysis for Summaries}\label{sec:abstraction}

Abstraction and Faithfulness are normally two opposing properties. For dataset-specific summaries, despite their novel $n$-grams being higher than element-aware summaries in many cases, they sacrifice factual correctness to some extent, which is a fake high-abstraction. Case comparisons are shown in Table \ref{tab:abstraction_faithfulness1}-\ref{tab:abstraction_faithfulness2}.





        


\section{Better Understanding Summary Chain-of-Thought: Case Study}\label{sec:casestudy}

\subsection{Learn How SumCoT Works}\label{sec:gpt3_vs_gpt3sumcot}

Table \ref{tab:gpt3_gpt3sumcot1}-\ref{tab:gpt3_gpt3sumcot3} presents some cases to 
visually show how SumCoT affects the final generated summary.
We compare GPT-3 zero-shot summaries before and after using SumCoT. Core elements have been highlighted in the table. It is clear that summaries using SumCoT Cover a large number of fine-grained elements extracted by GPT-3 in Stage 1 that are not in the standard zero-shot summaries.

\subsection{Error Analysis for Element Extractions}\label{sec:errorcase}

To validate the correctness of element extraction of LLMs, we conduct a large number of sampling observations, and summarize the two main issues:

\begin{itemize}[leftmargin=*]

\item \textbf{Date Hallucination.}  
This issue is mainly caused by two aspects: (1) \textbf{Date} elements are not presented in many cases, so this requires LLMs to question date existence rather than provide false dates, but LLMs are hardly aware of this situation.
(2) In more difficult cases, date extraction involves reasoning (e.g. "\textit{In 2014... Two years ago...}" $\rightarrow$ "\textit{In 2012}"), which poses a greater challenge for extraction, and causes the sometimes failure of LLMs. Cases are presented in Table \ref{tab:error1}-\ref{tab:error3},
these all explain why the $F_1$ score of \textbf{Date} is lower than that of the other elements (Table \ref{tab:extraction}).
\vspace{-0.1in}

\item \textbf{Element Redundancy.}  
LLMs frequently extract elements that are faithful to the source document but not important.
Cases are presented in Table \ref{tab:error4}-\ref{tab:error5}.
This explains why the $\mathrm{Precision}$ score is lower than the $\mathrm{Recall}$ score in almost every element (Table \ref{tab:extraction}).

\end{itemize}

\subsection{Ablation Study of GPT-3 Model Size}\label{sec:ablation}

We try diverse versions of GPT-3 with different model sizes. 
Model configurations are as follows:

\begin{itemize}
    \item 0.3B-parameter \texttt{text-ada-001}
    \vspace{-0.1in}
    \item 1.3B-parameter \texttt{text-babbage-001}
    \vspace{-0.1in}
    \item 6.7B-parameter \texttt{text-curie-001}
    \vspace{-0.1in}
    \item 175B-parameter \texttt{text-davinci-002}
\end{itemize}

The curve of $F_1$ score of different versions has been shown in Figure \ref{img:modelsize}, and the case study is presented in Table \ref{tab:ablation}.

\section{Random Sample Presentation}
\label{sec:randomsamples}
We randomly sample some examples, each containing: source document, golden summary, expert-writing summary, GPT-3 zero-shot summary, and GPT-3 reasoning-like zero-shot summary. Examples are shown in Table \ref{tab:randomsamples1}-\ref{tab:randomsamples4}.

\begin{table*}[t]
\centering
\footnotesize
  \renewcommand\arraystretch{1}
  \setlength{\tabcolsep}{2.5mm}{

\begin{tabular}{p{0.96\columnwidth}p{0.96\columnwidth}}
\toprule

\textbf{Source Document (\textit{CNN/DailyMail})} \\
\midrule

\multicolumn{2}{p{1.98\columnwidth}}
{
(The Hollywood Reporter) Add another fan-favorite character to the cast of next year's "X-Men: Apocalypse" with director Bryan Singer announcing via Instagram that Olivia Munn will play the telepathic Psylocke in the follow-up to "X-Men: Days of Future Past." Singer revealed that the "Newsroom" actress would play Betsy Braddock in the movie (presumably before the confusing and complicated plot twist that saw Psylocke change from a Caucasian former supermodel to a Japanese ninja for no immediately obvious reason). \"Apocalypse\" is currently in production for a summer 2016 release. More: "X-Men: Apocalypse" casts fan favorite Jubilee. The comic book's Psylocke was created by Chris Claremont and Herb Trimpe for the British "Captain Britain" series, where she appeared throughout the 1970s and '80s, before joining the X-Men in 1987's "Uncanny X-Men" No. 213. Since that time, she has been a mainstay both of the main team and spin-off series including "Exiles" and "X-Force." More: What newcomers need to know about Marvel's "Secret Wars". Munn will join a cast that includes James McAvoy, Michael Fassbender and Jennifer Lawrence in the movie, which hits theaters May 27, 2016. Munn is repped by Creative Artists Agency and Atlas Artists. More: Does the big plot twist in "Terminator Genisys" blow up the franchise? @The Hollywood Reporter. All rights reserved." 
}
\\
\midrule

\textbf{Dataset-specific Summary} & \textbf{Element-aware Summary} \\
\midrule

Olivia Munn will play Psylocke in "x-men: apocalypse" film. \textcolor{orange}{Psylocke trended for hours on twitter after director Bryan Singer announced casting}. 
& Olivia Munn will play the telepathic Psylocke -created by Chris Claremont and Herb Trimpe for the \"Captain Britain\" series - in the \"X-Men: Apocalypse\". The movie will be released in May 27, 2016. \\
\midrule
\textit{\% of novel uni/bi/trigram: 47.61/70.00/84.21} & \textit{\% of novel uni/bi/trigram: 17.65/48.51/62.50} \\

\bottomrule
\end{tabular}
\caption{Comparisons between element-aware summaries and dataset-specific summaries in abstraction and faithfulness. Hallucinatory facts are highlighted in \textcolor{orange}{orange}. We observe that dataset-specific summaries contain more hallucinatory facts despite a higher percentage of novel $n$-grams (Appendix \ref{sec:abstraction}).}
\label{tab:abstraction_faithfulness1}%
}

\end{table*}

\begin{table*}[t]
\centering
\footnotesize
  \renewcommand\arraystretch{1}
  \setlength{\tabcolsep}{2.5mm}{

\begin{tabular}{p{0.96\columnwidth}p{0.96\columnwidth}}

\toprule
\textbf{Source Document (\textit{BBC XSum})}\\
\midrule

\multicolumn{2}{p{1.98\columnwidth}}
{
More than 350 roma people had lived in the camp on la petite ceinture since mid-2015. Activists said many left early ahead of the police action. The site belongs to the national rail authority sncf. France has one of europe's toughest policies towards roma. Most live in camps that are regularly demolished and every year thousands are deported. Amnesty international urged city authorities to find a lasting housing solution for those evicted in paris - saying they would become homeless in mid-winter. Hundreds of thousands of roma - mostly from romania and bulgaria - have moved to western europe since the 1990s. The council of europe, the region 's main human rights body, warned that evictions were " counter-productive" because they disrupted education and healthcare for roma children. Council of europe secretary general thorbjorn jagland said it was crucial for french authorities to provide "adequate alternative accommodation" for those evicted, particularly as they have decided to take this action during winter.
}
\\
\midrule

\textbf{Dataset-specific Summary} & \textbf{Element-aware Summary} \\
\midrule
\textcolor{orange}{Police have cleared hundreds of roma people} from a slum-like camp built on a disused rail line in north paris. &  Every year thousands of Roma people are deported by France, and the region's main human rights body urges France to provide alternative accommodation for those evicted.
 \\
\midrule
\textit{\% of novel uni/bi/trigram: 28.57/85.00/100.00} & \textit{\% of novel uni/bi/trigram: 25.93/53.85/80.00} \\

\bottomrule      
        
\end{tabular}
}

\caption{Comparisons between element-aware summaries and dataset-specific summaries in abstraction and faithfulness. Hallucinatory facts are highlighted in \textcolor{orange}{orange}. We observe that dataset-specific summaries contain more hallucinatory facts despite a higher percentage of novel $n$-grams (Appendix \ref{sec:abstraction}).}
\label{tab:abstraction_faithfulness2}%

\end{table*}

\begin{table*}[t]
\centering
\footnotesize
  \renewcommand\arraystretch{1}
  \setlength{\tabcolsep}{2.5mm}{

\begin{tabular}{p{1.98\columnwidth}}

\toprule
\toprule
\textbf{Source Document (\textit{CNN/DailyMail})}\\
\midrule
Once famed for his mop of blacker than black hair, disgraced Democrat Rod Blagojevich, 58, has really let his haircare regime go while he serves his prison time. The former Illinois governor has return to his roots while inside and has been photographed with his still full head of hair a shocking white color rather than the boot polish black that was his trademark as a politician. Blagojevich was infamously caught trying to sell Barack Obama's U.S. Senate seat when he was elected president in 2008. Fade to gray: Once famed for his mop of blacker than black hair, disgraced Democrat Rod Blagojevich, 58, has really let his haircare regime go while he serves his prison time. Back in his days as governor of Illinois, Blagojevich was famed for his boot polish black hair. He was impeached and removed from office by the state Legislature in early 2009 following his arrest on federal corruption charges. Blagojevich is currently serving a 14-year sentence at the Federal Correctional Institution Englewood near Denver. The photos, the first of Blagojevich in prison, were first published by the National Enquirer. The photos show the former governor sitting at a desk, reading a book and wearing glasses and sitting outside in shorts and a wife beater shirt. While he continues to awaiting word on a last-ditch appeal, Blagojevich has daily contact with his family and teaches the history of war battles to other inmates, his attorney said. Last July lawyers attempted to argue that a high court ruling in McCutcheon v. the Federal Election Commission in April 2014 bolstered their view that Blagojevich was engaged in legal, run-of-the-mill political horse trading, not corruption. The McCutcheon ruling, they argued, found that soliciting contributions crosses the line into corruption only when a politician makes a clear, explicit promise to take official action in return for a donation. Blagojevich is currently serving a 14-year sentence at the Federal Correctional Institution Englewood near Denver after he was caught trying to sell Barack Obama's U.S. Senate seat when he was elected president. In that context, defense attorneys said the trial judge was wrong to set a lower standard for jurors by telling them, to convict, they only needed to find that Blagojevich sought donations 'knowing or believing' it would be given for some official act. But government attorneys responded saying that Blagojevich's legal team hadn't correctly characterized the Supreme Court's findings in McCutcheon. 'The decision provides no support for Blagojevich's argument on appeal,' the two-page filing said. 'Nothing in the (McCutcheon) decision suggests that an exchange of contributions for specific official acts is quid pro quo corruption only if the arrangement is stated 'explicitly' or 'expressly.' The Chicago-based appellate court has been considering Blagojevich's appeal for more than a year months. It's unclear when it might issue a decision. \\

\midrule
\midrule
\textbf{GPT-3 Zero-shot Summary (Standard)} \\
\midrule
\textit{Summarize the above article:} \\
The article discusses how \textcolor{red}{Rod Blagojevich}, a former Illinois governor, has let his hair go while \textcolor{blue}{serving his prison sentence}. The photos show the former governor with white hair, rather than the black hair that was his trademark as a politician. Blagojevich is currently serving \textcolor{green}{a 14-year sentence} at \textcolor{red}{the Federal Correctional Institution Englewood} near Denver. \\

\midrule
\midrule
\textbf{Summary Chain-of-Thought (Stage 1, Element Extraction)} \\
\midrule
\textit{What are the important entities in this document? What are the important dates in this document? What events are happening in this document? What is the result of these events?} \\

\textit{Please answer the above questions:} \\

1. The important entities in this document are \textcolor{red}{Rod Blagojevich}, \textcolor{red}{the Federal Correctional Institution Englewood}, and the National Enquirer. \\
2. The important dates in this document are \textcolor{yellow}{2008}, \textcolor{yellow}{2009}, and \textcolor{yellow}{2014}. \\
3. The events in this document are Blagojevich's \textcolor{blue}{arrest} and \textcolor{blue}{imprisonment}, and \textcolor{blue}{his appeal of his conviction}. \\
4. The result of these events is that Blagojevich is currently serving \textcolor{green}{a 14-year sentence in prison}. \\

\midrule
\textbf{GPT-3 Zero-shot-cot Summary (Stage 2, Summary Generation)} \\
\midrule
\textit{Let’s integrate the above information and summarize the article:} \\
In \textcolor{yellow}{2008}, \textcolor{red}{Rod Blagojevich} \textcolor{blue}{was arrested} for trying to sell Barack Obama's U.S. Senate seat. He was impeached and removed from office in \textcolor{yellow}{2009}, and \textcolor{green}{sentenced to 14 years in prison}. He is currently \textcolor{blue}{serving his sentence} at \textcolor{red}{the Federal Correctional Institution Englewood near Denver}. In \textcolor{yellow}{2014}, his lawyers attempted to \textcolor{blue}{appeal his conviction}, arguing that a recent Supreme Court ruling supported their view that Blagojevich was engaged in legal, run-of-the-mill political horse trading, not corruption.  \\

\bottomrule
\bottomrule

\end{tabular}
}

\caption{More comparisons between GPT-3 zero-shot summaries before and after using SumCoT. Spans of \textit{\textbf{Entity}}, \textit{\textbf{Date}}, \textit{\textbf{Event}}, and \textit{\textbf{Result}} are separately highlighted in \textcolor{red}{red}, \textcolor{yellow}{yellow}, \textcolor{blue}{blue} and \textcolor{green}{green}. \textbf{Prompts} are presented in \textit{italics} (Appendix \ref{sec:gpt3_vs_gpt3sumcot}).}
\label{tab:gpt3_gpt3sumcot1}%

\end{table*}

\begin{table*}[t]

\centering
\footnotesize
  \renewcommand\arraystretch{1}
  \setlength{\tabcolsep}{2.5mm}{

\begin{tabular}{p{1.98\columnwidth}}

\toprule
\toprule

\textbf{Source Document (\textit{BBC XSum})}\\
\midrule

The 69-year-old's yamaha collided with a Nissan car between handley's corner and barre garroo crossroads at about 17:00 bst on 4 June. Mr. Baker, who was from the island, was airlifted to noble's hospital, where he later died. Insp dave dobbie said "thoughts and condolences are with Mr. baker's family". He added that investigations are ongoing in relation to the crash. The car driver, who police say was northern irish, was treated in hospital but has been discharged. Another motorcyclist who was injured after the crash has also been released from hospital. The collision happened outside of the TT racing, which is taking place on the island.\\

\midrule
\midrule
\textbf{GPT-3 Zero-shot Summary (Standard)} \\
\midrule
\textit{Summarize the above article in one sentence:} \\
The article discusses \textcolor{blue}{a fatal motorcycle collision} that occurred on the Isle of Man, and \textcolor{blue}{the investigation} that is ongoing.\\

\midrule
\midrule
\textbf{Summary Chain-of-Thought (Stage 1, Element Extraction)} \\
\midrule
\textit{What are important entities in this document? What are important dates in this document? What events are happening in this document? What is the result of these events?} \\

\textit{Please answer the above questions:} \\

1. The important entities in this document are \textcolor{red}{Mr. Baker}, \textcolor{red}{the car driver}, and \textcolor{red}{the motorcyclist} who was injured. \\
2. The important dates in this document are \textcolor{yellow}{4 June} and the present day.\\
3. The events happening in this document are a \textcolor{blue}{collision between Mr. Baker's motorcycle and a car}, and \textcolor{blue}{the investigation into the collision}.\\
4. The result of these events is that Mr. Baker \textcolor{green}{died} and the car driver and motorcyclist were \textcolor{green}{injured}. \\

\midrule
\textbf{GPT-3 Zero-shot-cot Summary (Stage 2, Summary Generation)} \\
\midrule
\textit{Let’s integrate the above information and summarize the article in one sentence:} \\
\textcolor{yellow}{On 4 June}, \textcolor{red}{Mr. Baker}'s \textcolor{blue}{motorcycle collided with a car}, resulting in \textcolor{green}{his death}. The \textcolor{red}{car driver} and \textcolor{red}{motorcyclist} were \textcolor{green}{injured}. \\

\bottomrule 
\bottomrule 

\end{tabular}
}

\caption{More comparisons between GPT-3 zero-shot summaries before and after using SumCoT. Spans of \textit{\textbf{Entity}}, \textit{\textbf{Date}}, \textit{\textbf{Event}}, and \textit{\textbf{Result}} are separately highlighted in \textcolor{red}{red}, \textcolor{yellow}{yellow}, \textcolor{blue}{blue} and \textcolor{green}{green}. \textbf{Prompts} are presented in \textit{italics} (Appendix \ref{sec:gpt3_vs_gpt3sumcot}).}
\label{tab:gpt3_gpt3sumcot2}%
    
\end{table*}

\begin{table*}[t]

\centering
\footnotesize
  \renewcommand\arraystretch{1}
  \setlength{\tabcolsep}{2.5mm}{

\begin{tabular}{p{1.98\columnwidth}}

\toprule
\toprule
\textbf{Source Document (\textit{BBC XSum})}\\
\midrule

West midlands police said the man had been detained in reading and the woman in wokingham by counter-terrorism officers at about 19:30 bst. The pair, believed to be friends, were taken to a police station in the west midlands for questioning. The force said the arrests had been intelligence-led. the suspects did not pose any imminent threat, a spokesperson added. Supt robin rickard from thames valley police said the arrests "were part of an ongoing investigation by west midlands counter terrorism unit and were pre-planned". "Officers from reading and wokingham are providing support and there are searches taking place at the suspects' home addresses," he said. \\

\midrule
\midrule
\textbf{GPT-3 Zero-shot Summary (Standard)} \\
\midrule
\textit{Summarize the above article in one sentence:} \\
\textcolor{red}{Two people} were \textcolor{blue}{arrested} in England on suspicion of terrorism. \\

\midrule
\midrule
\textbf{Summary Chain-of-Thought (Stage 1, Element Extraction)} \\
\midrule
\textit{What are the important entities in this document? What are the important dates in this document? What events are happening in this document? What is the result of these events?} \\

\textit{Please answer the above questions:} \\

The important entities in this document are the \textcolor{red}{two suspects} who were arrested, \textcolor{red}{the police forces} who made the arrests, and \textcolor{red}{the locations where the arrests took place}. The important dates are \textcolor{yellow}{the day of the arrests (19:30 BST)} and the day of the document (20:00 BST). The events happening in this document are \textcolor{blue}{the arrests of the two suspects} and \textcolor{blue}{the searches of their home addresses}. The result of these events is that the two suspects are in police custody and their home addresses are being searched. \\

\midrule
\textbf{GPT-3 Zero-shot-cot Summary (Stage 2, Summary Generation)} \\
\midrule
\textit{Let’s integrate the above information and summarize the article in one sentence:} \\
\textcolor{yellow}{On 19:30 BST}, \textcolor{red}{two suspects} were \textcolor{blue}{arrested} by \textcolor{red}{police forces} in \textcolor{red}{Reading and Wokingham} on intelligence-led grounds, and \textcolor{blue}{their home addresses are currently being searched}. \\

\bottomrule 
\bottomrule 

\end{tabular}
}

\caption{More comparisons between GPT-3 zero-shot summaries before and after using SumCoT. Spans of \textit{\textbf{Entity}}, \textit{\textbf{Date}}, \textit{\textbf{Event}}, and \textit{\textbf{Result}} are separately highlighted in \textcolor{red}{red}, \textcolor{yellow}{yellow}, \textcolor{blue}{blue} and \textcolor{green}{green}. \textbf{Prompts} are presented in \textit{italics} (Appendix \ref{sec:gpt3_vs_gpt3sumcot}).}
\label{tab:gpt3_gpt3sumcot3}%
    
\end{table*}

\begin{table*}[t]
\centering
\footnotesize
  \renewcommand\arraystretch{1}
  \setlength{\tabcolsep}{2.5mm}{

\begin{tabular}{p{1.98\columnwidth}}

\toprule
\toprule
\textbf{Error Type: Date Hallucination} \\
\midrule
\midrule

\textbf{Source Document (\textit{CNN/DailyMail})}\\
\midrule

Charity runners taking part in a 10km fun run at the weekend were left exhausted after being sent on an unscheduled two-mile detour. The blunder was believed to have been caused by a race marshal taking a toilet break during the event, missing 300 runners who should have been directed at a junction point. Instead they continued past the unmanned marshall point and had to run for an extra three kilometres while the other 900 competitors followed the correct route. Scroll down for video Blunder: Charity runners taking part in yesterday's Bournemouth Bay 10K Run (pictured) were left exhausted after being sent on an unscheduled two-mile detour. The bizarre gaffe happened during yesterday's Bournemouth Bay Run and today the organisers - Bournemouth Borough Council - appealed for those who were affected by the mix-up to contact them for a 'gesture of goodwill.'A local authority spokesman said that it was investigating what happened to the marshal who should have directed runners at a turning point. It was reported that some runners were 'in tears' while one described the event's organisation as 'shambolic'. Hayley James, who is four months pregnant and from Poole, said: 'To have a race of that scale with only one marshal on a point is inexcusable.'We saw loads of people walking at the end, some were in tears, I felt so sorry for them - I felt like crying at the 10km mark.'Andy Isaac, from Bournemouth, said the event was 'mayhem' with one point where an elderly woman managed to drive onto the route and was flashing her lights at oncoming runners. A map shows where up to 300 runners continued along the coastal path after a marshal who was meant to direct them on to a turn went to the toilet. Reaction: Two people vent their frustration at the Bournemouth Bay Run on Twitter yesterday. It also emerged that water stations ran out of supplies during the race, forcing some runners to drink from half-empty bottles that had been left on the ground as they battled against dehydration.Commenting on the Daily Echo website, one runner said: "We had a bottle of water at the three mile station, but at the six mile point they had totally ran out, so nothing. Jon Weaver, head of resort marketing and events at the council, said: 'Unfortunately there was some confusion with marshalling arrangements at one point, but it was a critical point. We apologise unreservedly to those front runners.'In 33 years of running... this is the first time this has happened and as part of our debrief we will be analysing the arrangements carefully... to learn for 2016.'We understand runners have trained for a long time for the event and it's hard for them and we do empathise with how they are feeling.'It was hoped that the event would have raised more than £70,000 for the British Heart Foundation. Some racers took to Twitter to vent their frustration over the blunder. Rob Kelly wrote: Really disappointed in the \#BournemouthBayRun 10k that ended up 13k very poor show bad marshalling \#wontbeback.' And Kim Kelly replied: 'Totally agree and never got to do a 5k as they were 45mins behind schedule :(((.' \\
\textit{What are the important entities in this document?}\\
\textit{What are the important dates in this document?}\\
\textit{What events are happening in this document?}\\
\textit{What is the result of these events?}\\
\textit{Please answer the above questions:} \\

\midrule
\textbf{Element Extraction}\\
\midrule

1. The important entities in this document are the charity runners, the race marshal, and the Bournemouth Borough Council. \\
2. The important dates in this document are \textcolor{blue}{the date of the race (May 3, 2015)} and \textcolor{blue}{the date the runners were supposed to be directed at the turning point (May 2, 2015)}.\\
3. The events happening in this document are the charity runners being sent on an unscheduled two-mile detour, and the race marshal taking a toilet break.\\
4. The result of these events is that the charity runners were left exhausted after being sent on an unscheduled two-mile detour.\\

{\color{red}{\XSolidBrush}} {\color{red}{(\textbf{The dates highlighted in \textcolor{blue}{blue} are not declared in the source document.})}} \\

\bottomrule

\end{tabular}
}

\caption{Element extraction errors of GPT-3. \textbf{Prompts} are presented in \textit{italics} (Appendix \ref{sec:errorcase}).}
\label{tab:error1}%

\end{table*}

\begin{table*}[t]
\centering
\footnotesize
  \renewcommand\arraystretch{1}
  \setlength{\tabcolsep}{2.5mm}{

\begin{tabular}{p{1.98\columnwidth}}

\toprule
\toprule
\textbf{Error Type: Date Hallucination} \\
\midrule
\midrule

\textbf{Source Document (\textit{BBC XSum})}\\
\midrule

patrick Bauer and Nathan Byrne helped Karl Robinson's men complete a perfect smash and grab, cancelling out Zach clough's free-kick. a victory which furthers the Addicks' own play-off ambitions was made all the more impressive considering that they were a man short from the 12th minute onwards following lewis page's red card. after page was sent off for a professional foul on josh vela, clough then made the punishment count double with a brilliant free-kick, his ninth goal of the season. but robinson 's side refused to lie down, equalising on 23 minutes when jake forster-caskey 's free-kick was turned in at the far post by defender bauer. the Addicks then grabbed a shock lead in first-half stoppage time when Forster-Caskey led an impressive counter, feeding Byrne to slot the ball under the advancing Ben Alnwick. bolton, now without a win in their last five games in all competitions, threw all they had at Charlton after that and struck the woodwork when Andrew Taylor's cross deflected off Chris Solly. they failed to seriously test goalkeeper Declan rudd in the second half, however, as they slipped to a second consecutive 2-1 defeat in front of their own fans. Charlton, unbeaten since christmas, have now picked up 11 points from their last five matches to climb to 11th, just six points off a play-off place. report supplied by the press association. \\
\textit{What are the important entities in this document?}\\
\textit{What are the important dates in this document?}\\
\textit{What events are happening in this document?}\\
\textit{What is the result of these events?}\\
\textit{Please answer the above questions:} \\

\midrule
\textbf{Element Extraction}\\
\midrule

1. The important entities in this document are Charlton Athletic, Karl Robinson, Zach Clough, and Ben Alnwick.\\
2. The important dates in this document are \textcolor{blue}{December 25th} and \textcolor{blue}{January 23rd}.\\
3. The events happening in this document are a football match between Charlton Athletic and Bolton Wanderers, and the aftermath of the match.\\
4. The result of these events is that Charlton Athletic won the match 2-1.\\

{\color{red}{\XSolidBrush}} {\color{red}{(\textbf{The dates highlighted in \textcolor{blue}{blue} are not declared in the source document.})}} \\

\bottomrule

\end{tabular}
}

\caption{Element extraction errors of GPT-3. \textbf{Prompts} are presented in \textit{italics} (Appendix \ref{sec:errorcase}).}
\label{tab:error2}%

\end{table*}

\begin{table*}[t]
\centering
\footnotesize
  \renewcommand\arraystretch{1}
  \setlength{\tabcolsep}{2.5mm}{

\begin{tabular}{p{1.98\columnwidth}}

\toprule
\toprule
\textbf{Error Type: Date Hallucination} \\
\midrule
\midrule

\textbf{Source Document (\textit{BBC XSum})}\\
\midrule

club director Emilio butragueno confirmed FIFA had asked the Spanish federation for documents relating to transfers over the last five years. but ragueno said the club was `` absolutely calm '' about its behaviour. \textbf{fifa handed Barcelona a transfer ban until 2016 after they breached regulations when signing 10 players under the age of 18 in 2013 .} in a statement, real Madrid said the majority of the 51 players that world football's governing body had asked for information on were exempt from regulations for various reasons. FIFA rules state that international transfers are only permitted for players who are over the age of 18 - unless the player in question meets one of three qualifying criteria. under-18s can move to a club in a different country if: real claim 10 of the 51 players were Spanish nationals, while 23 foreign players had already been registered with a Spanish club. on Thursday, real completed the \# 2.3 m signing of 16-year-old Norway midfielder Martin Odegaard from stromgodset. \textbf{barcelona are not allowed to sign any players in 2015 after their appeal against their transfer ban was dismissed in December.}\\
\textit{What are the important entities in this document?}\\
\textit{What are the important dates in this document?}\\
\textit{What events are happening in this document?}\\
\textit{What is the result of these events?}\\
\textit{Please answer the above questions:} \\

\midrule
\textbf{Element Extraction}\\
\midrule

1. The important entities in this document are FIFA, Barcelona, and Real Madrid.\\
2. The important dates in this document are 2013, when Barcelona was first given a transfer ban, and \textcolor{blue}{December of 2014}, when Barcelona's appeal against the ban was denied.\\
3. The events in this document are FIFA's investigation into Barcelona's transfer practices, and the eventual imposition of a transfer ban on the club.\\
4. The result of these events is that Barcelona is not allowed to sign any players in 2015.\\

{\color{red}{\XSolidBrush}} {\color{red}{(\textbf{The date when Barcelona's appeal against the ban was denied is likely to be December 2013 or December 2014 deduced from the bolded part of the source document, so the extracted date highlighted in \textcolor{blue}{blue} is evidence-insufficient.})}} \\

\bottomrule

\end{tabular}
}

\caption{Element extraction errors of GPT-3. \textbf{Prompts} are presented in \textit{italics} (Appendix \ref{sec:errorcase}).}
\label{tab:error3}%

\end{table*}

\begin{table*}[t]
\centering
\footnotesize
  \renewcommand\arraystretch{1}
  \setlength{\tabcolsep}{2.5mm}{

\begin{tabular}{p{1.98\columnwidth}}

\toprule
\toprule
\textbf{Error Type: Element Redundancy} \\
\midrule
\midrule

\textbf{Source Document (\textit{BBC XSum})}\\
\midrule

there were concerns about the stability of the building, which has been derelict for years, after the collapse of an internal wall last week. contractors began work on Friday night. the environment minister Alex Atwood recommended the demolition in the interests of public safety. mary o'dwyer, chair of Foyle civic trust, said a strategic approach is needed for the future protection of heritage in the city following the demolition . `` many of our members believe that this is a very sad event . `` the factory is part of our industrial heritage and more should have been done to save it . `` others believe something had to be done because it was a health and safety issue, '' she said . `` we would like to see a consensus and a strategic vision of all of the organizations in the city that have a remit for protecting our heritage. '' Foyle road, which was closed between the lower deck of the Craigavon bridge and john street roundabout, has reopened.
\\
\textit{What are the important entities in this document?}\\
\textit{What are the important dates in this document?}\\
\textit{What events are happening in this document?}\\
\textit{What is the result of these events?}\\
\textit{Please answer the above questions:} \\

\midrule
\textbf{Element Extraction}\\
\midrule

The important entities in this document are \textcolor{blue}{the Foyle Civic Trust}, the environment minister Alex Atwood, and \textcolor{blue}{the contractors who began work on Friday night}. The important dates in this document are the collapse of an internal wall last week, and the demolition of the factory on Friday night. The events happening in this document are the demolition of the factory, and the reopening of Foyle Road. The result of these events is that \textcolor{blue}{the factory is gone} and Foyle Road is open again.\\

{\color{red}{\XSolidBrush}} {\color{red}{(\textbf{The elements highlighted in \textcolor{blue}{blue} is redundent.})}} \\

\bottomrule

\end{tabular}
}

\caption{Element extraction errors of GPT-3. \textbf{Prompts} are presented in \textit{italics} (Appendix \ref{sec:errorcase}).}
\label{tab:error4}%

\end{table*}

\begin{table*}[t]
\centering
\footnotesize
  \renewcommand\arraystretch{1}
  \setlength{\tabcolsep}{2.5mm}{

\begin{tabular}{p{1.98\columnwidth}}

\toprule
\toprule
\textbf{Error Type: Element Redundancy} \\
\midrule
\midrule

\textbf{Source Document (\textit{BBC XSum})}\\
\midrule
some of those residents had only just returned to their homes after last week's eruptions. the latest eruption was less powerful but sent a large plume of dark grey smoke and ash rising from the crater. calbuco surprised residents of the Los Lagos region last week by bursting into life after decades of inactivity. the national geology and mines service issued a red alert after Thursday's eruption, warning that the volcano remained unstable. on Friday, residents were warned that heavy rain could mix with ash and rock to produce deadly volcanic mudflows. these `` lahars '' can slide down volcanoes like avalanches. the BBC's Gideon long in Santiago says that whereas the ash last week was blown in a north-easterly direction, this time it is drifting south-east over the town of cochamo. the town has been blanketed in a cloud of smoke and ash. calbuco is one of the most active volcanoes in chile, having erupted in 1961 and 1972, and many times before that. last week's eruptions caused the cancellation of flights to and within chile and forced farmers in the area to evacuate their livestock.
\\
\textit{What are the important entities in this document?}\\
\textit{What are the important dates in this document?}\\
\textit{What events are happening in this document?}\\
\textit{What is the result of these events?}\\
\textit{Please answer the above questions:} \\

\midrule
\textbf{Element Extraction}\\
\midrule

1. The important entities in this document are the volcano, the residents, and \textcolor{blue}{the government}.\\
2. The important dates in this document are \textcolor{blue}{the dates of the eruptions}.\\
3. The events happening in this document are the eruptions of the volcano and \textcolor{blue}{the resulting evacuations}.\\
4. The result of these events is that the residents have been forced to leave their homes and the area around the volcano is covered in ash.\\

{\color{red}{\XSolidBrush}} {\color{red}{(\textbf{The elements highlighted in \textcolor{blue}{blue} is redundent.})}} \\

\bottomrule

\end{tabular}
}

\caption{Element extraction errors of GPT-3. \textbf{Prompts} are presented in \textit{italics} (Appendix \ref{sec:errorcase}).}
\label{tab:error5}%

\end{table*}

\begin{table*}[t]
\centering
\footnotesize
  \renewcommand\arraystretch{1}
  \setlength{\tabcolsep}{2.5mm}{
\begin{tabular}{p{450pt}}
\toprule

\textbf{Source Document (\textit{BBC XSum})} \\
\midrule
They say every cloud has a silver lining. At least that's what one Sydney woman, caught out in this week's super storm, is hoping. With the rain clearing, a woman known only by her email address as MissieSmile21, has taped posters to the city's light posts to find the 'tall, charming gentleman' who offered her his umbrella to cross the street.Scroll down for video  The 'Love Call' notes are posted at traffic lights on Sydney's George Street near Australia SquareActively looking: Missie Smile is hoping the generous stranger will get in contact. The 'Love Call' notices, posted on George Street at Australia Square in Sydney's CBD, are appealing to the man with 'the most beautiful smile' who was in the area about '7:10 pm-ish' on Tuesday evening.'I am actively looking for you,' the poster reads. 'If you're single, let's grab a drink together.' People struggled against the heavy rain and strong winds in the Sydney CBD as the storm lashed NSW on Tuesday this weekMost people gave up on their umbrellas in winds of more than 100 km/ hr. \\


\textit{What are the important entities in this document?}

\textit{What are the important dates in this document?}

\textit{What events are happening in this document?}

\textit{What is the result of these events?}

\textit{Please answer the above questions:} \\

\midrule
\textbf{\texttt{text-ada-001}} \\
\midrule

The 'Love Call' notes are posted at traffic lights on Sydney's George Street near Australia Square.

The 'Love Call' notices, posted on George Street at Australia Square, are appealing to the man with 'the most beautiful smile' who was in the area about '7:10 pm-ish' on Tuesday evening.

People struggled against the heavy rain and strong winds in the Sydney CBD as the storm lashed NSW on Tuesday this week.

Most people gave up on their umbrellas in winds of more than 100 km/ hr.

What are important dates in this document?

What are important entities in this document?

What are important events in this document?

Please answer the above questions:

The 'Love Call' notes are posted at traffic lights on Sydney's George Street near Australia Square.

The 'Love Call' notices, posted on George Street at Australia Square, are appealing to the man with 'the most beautiful smile' who was in the area about '7:10 pm-ish' on Tuesday evening.

People struggled against the heavy rain and strong winds in the Sydney CBD as the storm lash NSW on Tuesday this week.

Most people gave up on their umbrellas in winds of more than 100 km/ hr.

What is the result of these events?

Please answer the above questions:

At 7:10 pm on Tuesday evening, a man with the most beautiful smile, known only as MissieSmile21, was in the area near Australia Square.




The 'Love Call' notices, posted on George Street at Australia Square, are appealing to the man with the most beautiful smile who was in the area about '7:10 pm-ish' on Tuesday evening.

People struggled against the heavy rain and strong winds in the Sydney CBD as the storm lash NSW on Tuesday this week.

Most people gave up on their umbrellas in winds of more than 100 km/ hr. \\

{\color{red}{\XSolidBrush}} {\color{red}{\textbf{(The extracted information is very confusing and illogical.)}}} \\

\midrule
\textbf{\texttt{text-babbage-001}} \\
\midrule

1. What is the email address of the woman whoposted the 'Love Call' notices on Sydney's George Street near Australia Square?

2. What is the date of the 'Love Call' notices?

3. What is the time of the 'Love Call' notices?

4. What is the type of person who was in the area around 7:10 pm-ish on Tuesday evening?

5. What is the result of the 'Love Call' notices? \\

{\color{red}{\XSolidBrush}} {\color{red}{\textbf{(The model fails to accurately understand the meaning of guiding questions.)}}} \\

\midrule
\textbf{\texttt{text-curie-001}} \\
\midrule

The important entities in this document are {\color{blue}{the poster}}, the man she is looking for, and the date {\color{blue}{7/10/2016}}. The important dates in this document are Tuesday, {\color{blue}{July 10th}}. The events in this document are the heavy rain and strong winds. The result of these events is that {\color{blue}{many people were unable to use their umbrellas and many people were stranded}}. \\

{\color{red}{\XSolidBrush}} {\color{red}{\textbf{(The model understands the guiding questions, but extracts many redundant or false facts shown in}}} {\color{blue}{\textbf{blue}}} {\color{red}{\textbf{spans.)}}} \\

\midrule
\textbf{\texttt{text-davinci-002}} \\
\midrule

The important entities in this document are MissieSmile21, the tall, charming gentleman, and George Street at Australia Square. The important dates are Tuesday evening and 7:10 pm. The events are the super storm and MissieSmile21's search for the tall, charming gentleman. The result of these events is unknown. \\

{\color{red}{\Checkmark}} {\color{red}{\textbf{(The model can accurately extract human-identified core elements from the source document.)}}} \\

\bottomrule      
        
\end{tabular}
    \caption{Element extraction results of different GPT-3 versions. \textbf{Prompts} are presented in \textit{italics} (Appendix \ref{sec:ablation}).}
    \label{tab:ablation}%
}
    
\end{table*}

\begin{table*}[t]
\centering
\footnotesize
  \renewcommand\arraystretch{1}
  \setlength{\tabcolsep}{2.5mm}{

\begin{tabular}{p{1.98\columnwidth}}

\toprule
\textbf{Source Document (\textit{CNN/DailyMail})}\\
\midrule

A Tory councillor with a history of odd behaviour was told to put down his knife and fork after being caught tucking into a roast duck dinner during a council meeting. Jonny Bucknell, 58, was enjoying his meal in the council chamber when a Labour rival, Theo Blackwell, spotted him and alerted other councillors. He was forced to put down his cutlery when the mayor, Lazzaro Pietragnoli, interrupted the proceedings to tell him off. Taking a stand: Jonny Bucknell is no stranger to odd behaviour. In 2013 he slept in his car at the Tory party conference. He now says he wants a rule change so he can eat a roast dinner at council meetings. The mayor, who was chairing the meeting of Camden Council in north London, reminded the hungry councillor that eating was banned in the chamber. But the angry diner claims he was unaware eating there was forbidden and said he now aims to campaign for a rule change. The rumpus comes a month after Liberal Democrat councillor Martin Elengorn was caught playing Scrabble during a Richmond Council budget meeting in south-west London. Telling off: Mayor of Camden Council, Lazzaro Pietragnoli, had to tell Mr Bucknell to stop eating. When he first noticed him eating, Mr Blackwell told his fellow councillors: 'It appears that one of our Tory colleagues is consuming a full Sunday roast dinner in the council chamber. 'Could I ask the borough solicitor to give us advice on eating a full roast dinner in the council chamber? It's a little bit more than a cheeky Snickers.' The diner was forced to curtail his meal. Mr Bucknell, who has been a councillor for more than ten years and represents Belsize, Hampstead, told the Evening Standard: 'I never knew there was a ban on eating in the chamber. 'They should definitely repeal it. There is nothing wrong with nibbling from a lunch box if you are being discreet. 'It is not as if a cold meal is going to waft around like a McDonald's. 'I will be campaigning for the repealing of the law that says you can't nibble from a lunch box in the council chamber.' The Conservative councillor said the meal, in a plastic box, had travelled home with him after a French snowboarding holiday. 'The chalet always brought out too much food and I can't stand wasting food,' he said. He previously found fame when he slept in his Volvo car to save on hotel bills during the Conservative party conference in September 2013. Mr Bucknell said at the time it was to make a stand against what he called 'ridiculous prices'. He said the economy would improve if more people were thrifty like him. After the council meeting mayor Lazzaro Pietragnoli said: 'I understand councillors do a difficult job and sometimes don't get time to eat dinner. 'I also tend to be quite flexible, but having a big meal in front of him - that was a bit too much.' A Camden Council spokesman said: 'It is as the mayor said, standing orders say that members should not eat in the chamber.' No eating: The Camden Council chamber where Councillor Bucknell was forced to curtail his roast dinner. \\

\midrule
\textbf{Dataset-specifc Summary}\\
\midrule

Labour rival alerted colleagues after spotting Councillor tucking into roast. It is not the first time jonny Bucknell, 58, has demonstrated odd behaviour.. In 2013 he slept in his car to make a point while attending Tory conference. Mr Bucknell said he will campaign for rule change about eating at meetings. \\

\midrule
\textbf{Element-aware Summary}\\
\midrule

Jonny Bucknell, a Tory councillor, was found tucking into a roast duck dinner during a council meeting. He now wants a rule change to have dinner at council meetings. It is not the first time for his strange behavior. In September 2013, Mr Bucknell slept in his Volvo car to save on hotel bills during the Conservative party conference to make a stand against what he called 'ridiculous prices'. \\

\midrule
\textbf{GPT-3 Zero-shot Summary}\\
\midrule

A Tory councillor was told to stop eating a roast duck dinner during a council meeting after a Labour rival spotted him and alerted other councillors. The councillor, Jonny Bucknell, claims he was unaware that eating in the council chamber was forbidden and said he now aims to campaign for a rule change. \\

\midrule
\textbf{GPT-3 Zero-shot-cot Summary}\\
\midrule

In September 2013, Councillor Jonny Bucknell made headlines when he slept in his car during the Conservative party conference in order to save on hotel bills. Recently, he has found himself in the news again after being caught eating a roast duck dinner during a council meeting. Councillor Bucknell claims that he was unaware that eating in the council chamber was forbidden, and is now campaigning for a rule change so that he can eat during meetings. \\

\bottomrule

\end{tabular}
}

\caption{Random samples from \textit{CNN/DailyMail} and \textit{BBC XSum} datasets (Appendix \ref{sec:randomsamples}).}
\label{tab:randomsamples1}%

\end{table*}

\begin{table*}[t]

\centering
\footnotesize
  \renewcommand\arraystretch{1}
  \setlength{\tabcolsep}{2.5mm}{

\begin{tabular}{p{1.98\columnwidth}}

\toprule
\textbf{Source Document (\textit{CNN/DailyMail})}\\
\midrule

This is the dramatic moments armed police swoop on a villa where a Briton linked to the gangland murder of a torture victim was arrested. Paul Monk, 54, from Essex, was wanted by Spanish police for questioning over the kidnap and murder of Francis Brennan, whose badly decomposed body washed up on a Costa Blanca beach in March last year. He was also wanted by the Metropolitan Police on drug offences and had been named on a list of fugitives published as part of the National Crime Agency's Operation Captura campaign ahead of his detention. This is the dramatic moment that fugitive Paul Monk was arrested by heavily armed police in his Alicante villa. Paul Monk, 54, from Essex, was wanted by Spanish police for questioning over the kidnap and murder of Francis Brennan. Spanish police released footage of their dramatic swoop. This grab for the video shows them approaching the villa at speed. The police move steathily up the steps of Monk's villa, weapons drawn. Taking no chances: The highly trained, well-armed police moved through the house room by room. Paul Monk was on the UK's most wanted list on suspicion of drug trafficking. Brennan, 25, from Liverpool, vanished in the resort of Javea in January last year after being kidnapped by men posing as police. His body was wrapped in an industrial-size bin bag with duct tape round it when it appeared on a beach in nearby Orihuela Costa. Civil Guard officers in Alicante confirmed today they believe Monk, from Essex, may be implicated in the violent death and named him as an associate of Paul Scott. Scott, 32, was arrested on a charge of conspiracy to import cocaine after being caught trying to sneak into Britain in a light aircraft last December. He was also wanted for questioning over Mr Brennan's murder when he was detained. Guardia Civil described him last night as the suspected mastermind of the crime. Monk was detained at a four-bedroom property in Javea near Alicante as he directed workers laying a marble patio around his swimming pool. An imitation firearm with a silencer and nearly \u00a3100,000 in cash were also found. He is being held in jail and is expected to be charged and face trial in Spain over Mr Brennan's murder before being extradited to the UK to face questioning over alleged drugs offences. He has been linked to the handover of one kilo of cocaine in Cockfosters, London, in May 2013 and the seizure of 24 kilos of cannabis in Colchester in October 2013. A Civil Guard spokesman said: 'He never left his house as a security measure to avoid being arrested. 'He got other people to bring him food and other things in the villa where he hid out, leading the life of an authentic fugitive.' The police raid had air support, with this grab coming from footage of Monk's villa taken by a helicopter. Wads of money found by armed police after they arrested Monk . Monk is being held in jail and is expected to be charged and face trial in Spain over Mr Brennan's murder before being extradited to the UK to face questioning over alleged drugs offences. Spanish police search Monk's property thoroughly for evidence, finding an imitation gun with a silencer. National Crime Agency detectives took part in the raid on the property along with Civil Guard officers. Mr Brennan, from Liverpool, was himself on the run after leaving Britain while awaiting sentence in the UK over the wounding of a man at a Swedish House Mafia music concert in Milton Keynes, Bedfordshire. He travelled to the Costa del Sol with his girlfriend towards the end of 2013 after ignoring the advice of his parents to stay in Britain. Police swooped on Scott last year after he flew from Holland to a remote airstrip in Norfolk. He had fake Belgian identity documents on him and more than \u00a31,500 in cash. He was jailed in January for plotting to import 40 tonnes of cocaine into Britain. Police described him as the 'last man outstanding in a criminal gang that plotted to smuggle vast quantities of cocaine into the UK' after he was sentenced to 14 years in prison at Liverpool Crown Court. Sorry we are not currently accepting comments on this article. \\

\midrule
\textbf{Dataset-specific Summary}\\
\midrule

Paul Monk, 54, was wanted by Spanish Police in connection with a murder.. The Essex man is a suspect in the murder of Francis Brennan.. Brennan's body washed up on a Costa Blanca beach in March last year.. Police released footage of their swoop on Monk's Alicante villa. \\

\midrule
\textbf{Element-aware Summary}\\
\midrule

Paul Monk, 54, from Essex, was wanted by Spanish police for questioning over the kidnap and murder of Francis Brennan, 25, from Liverpool. Brennan\u2019 body was found on a beach in March last year. Paul Monk was swoop by armed police on a villa, he will be arrested and face trial in Spain for this murder. \\

\midrule
\textbf{GPT-3 Zero-shot Summary}\\
\midrule

A man wanted for questioning in connection with the murder of a torture victim has been arrested by Spanish police. Paul Monk, 54, from Essex, was detained at his villa in Javea near Alicante as he directed workers laying a marble patio around his swimming pool. An imitation firearm with a silencer and nearly \u00a3100,000 in cash were also found. Monk is being held in jail and is expected to be charged and face trial in Spain over the murder before being extradited to the UK to face questioning over alleged drugs offences. \\

\midrule
\textbf{GPT-3 Zero-shot-cot Summary}\\
\midrule

Paul Monk, 54, from Essex, was arrested by Spanish police in his Alicante villa for questioning over the kidnap and murder of Francis Brennan, 25, from Liverpool. Brennan's body was found in March last year, wrapped in an industrial-size bin bag with duct tape round it, on a beach in nearby Orihuela Costa. Monk is being held in jail and is expected to be charged and face trial in Spain over Mr Brennan's murder before being extradited to the UK to face questioning over alleged drugs offences. \\

\bottomrule 

\end{tabular}}

\caption{Random samples from \textit{CNN/DailyMail} and \textit{BBC XSum} datasets (Appendix \ref{sec:randomsamples}).}
\label{tab:randomsamples2}%

\end{table*}

\begin{table*}[t]

\centering
\footnotesize
  \renewcommand\arraystretch{1}
  \setlength{\tabcolsep}{2.5mm}{

\begin{tabular}{p{1.98\columnwidth}}

\toprule
\textbf{Source Document (\textit{BBC XSum})}\\
\midrule

They've kicked up a stink with their council, who've decided to use the liquid to kill weeds around the city. It's all part of a trial to get rid of the unwanted plants without using strong chemicals, which some people think aren't environmentally friendly. But many locals say they're not happy with the unpleasant pong the trial's creating. They've started a campaign and also say they don't think vinegar is strong enough to kill the weeds. Bristol city council say they'll continue with their trial, but will keeping track of feedback from the people who live there. \\

\midrule
\textbf{Dataset-specific Summary}\\
\midrule

People living in bristol have complained about a vinegary whiff in the air. \\

\midrule
\textbf{Element-aware Summary}\\
\midrule

Bristol city is conducting a trial to kill weeds with vinegar, but the unpleasant pong makes the locals unhappy, which led to a conflict between the two. \\

\midrule
\textbf{GPT-3 Zero-shot Summary}\\
\midrule

The city of Bristol is conducting a trial to see if vinegar can be used as an environmentally friendly weed killer, but many locals are not happy with the strong smell. \\

\midrule
\textbf{GPT-3 Zero-shot-cot Summary}\\
\midrule

The city council of Bristol is conducting a trial to kill weeds using vinegar, but many locals are unhappy with the unpleasant smell the trial is creating. \\

\bottomrule 

\end{tabular}}

\caption{Random samples from \textit{CNN/DailyMail} and \textit{BBC XSum} datasets (Appendix \ref{sec:randomsamples}).}
\label{tab:randomsamples3}%

\end{table*}

\begin{table*}[t]

\centering
\footnotesize
  \renewcommand\arraystretch{1}
  \setlength{\tabcolsep}{2.5mm}{

\begin{tabular}{p{1.98\columnwidth}}

\toprule
\textbf{Source Document (\textit{BBC XSum})}\\
\midrule

Cpl Kylo gorman's medals for service in Afghanistan and at the queen 's diamond jubilee were taken from his car in bolton-le-sands on 4 or 5 January. He said he wasn't that bothered at first, but the reaction online got me thinking "it is a really big deal" . Lancashire police said they were investigating the theft. The port talbot-born 28-year-old, who has served five tours of Afghanistan, said he had left the medals in the car after removing them from a uniform he was having altered. "I didn't think twice about them. I'm based at raf leeming in north yorkshire during the week and when I came home and the car had been broken into, I realised my medals had gone," he said. "I want my children to be able to look at their dad's medals and be proud of what I did," he added. Cpl gorman, who has also served in the Falklands, Africa and Cyprus, has been in the RAF for 10 years and is due to leave the armed forces in April. \\

\midrule
\textbf{Dataset-specific Summary}\\
\midrule

A RAF airman has made an appeal for the return of two stolen medals after hundreds shared his post about the theft on social media. \\

\midrule
\textbf{Element-aware Summary}\\
\midrule

Cpl Gorman's medals were stolen on 4 or 5 january, and he has made an appeal for the return. Lancashire police are investigating the theft. \\

\midrule
\textbf{GPT-3 Zero-shot Summary}\\
\midrule

The medals of a British serviceman were stolen from his car, and he is now appealing for their return. \\

\midrule
\textbf{GPT-3 Zero-shot-cot Summary}\\
\midrule

Cpl Gorman's medals for service in Afghanistan and at the Queen's Diamond Jubilee were stolen from his car on 4 or 5 January, and Lancashire police are investigating the theft. \\

\bottomrule 

\end{tabular}}

\caption{Random samples from \textit{CNN/DailyMail} and \textit{BBC XSum} datasets (Appendix \ref{sec:randomsamples}).}
\label{tab:randomsamples4}%

\end{table*}

\end{CJK*}
\end{document}